\documentclass[sigconf]{acmart}

\usepackage{graphicx}
\usepackage{color}
\usepackage{dsfont}
\usepackage{mathrsfs}
\usepackage{amsmath}
\usepackage{multirow}
\usepackage{lscape}
\usepackage{subcaption}
\usepackage{enumitem}

\newcommand{\DEE}{\textit{Dual Emotion Features}}

\AtBeginDocument{%
  \providecommand\BibTeX{{%
    \normalfont B\kern-0.5em{\scshape i\kern-0.25em b}\kern-0.8em\TeX}}}

\copyrightyear{2021}
\acmYear{2021}
\setcopyright{iw3c2w3}
\acmConference[WWW '21]{Proceedings of the Web Conference 2021}{April 19--23, 2021}{Ljubljana, Slovenia}
\acmBooktitle{Proceedings of the Web Conference 2021 (WWW '21), April 19--23, 2021, Ljubljana, Slovenia}
\acmPrice{}
\acmDOI{10.1145/3442381.3450004}
\acmISBN{978-1-4503-8312-7/21/04}
\begin{document}

\title{Mining Dual Emotion for Fake News Detection}

\author{Xueyao Zhang}
\authornotemark[2]
\affiliation{%
  \institution{Institute of Computing Technology, Chinese Academy of Sciences}
  \institution{University of Chinese Academy of Sciences}
}
\email{zhangxueyao19s@ict.ac.cn}

\author{Juan Cao}
\authornote{Corresponding authors}
\authornote{At Key Laboratory of Intelligent Information Processing of Chinese Academy of Sciences. Also at State Key Laboratory of Communication Content Cognition, People's Daily Online.}
\affiliation{%
  \institution{Institute of Computing Technology, Chinese Academy of Sciences}
  \institution{University of Chinese Academy of Sciences}
}
\email{caojuan@ict.ac.cn}

\author{Xirong Li}
\authornotemark[1]
\affiliation{%
  \institution{Key Lab of Data Engineering and Knowledge Engineering, Renmin University of China}
  \city{Beijing}
  \country{China}
}
\email{xirong@ruc.edu.cn}

\author{Qiang Sheng}
\authornotemark[2]
\affiliation{%
  \institution{Institute of Computing Technology, Chinese Academy of Sciences}
  \institution{University of Chinese Academy of Sciences}
}
\email{shengqiang18z@ict.ac.cn}

\author{Lei Zhong}
\authornotemark[2]
\affiliation{%
  \institution{Institute of Computing Technology, Chinese Academy of Sciences}
  \institution{University of Chinese Academy of Sciences}
}
\email{zhonglei18s@ict.ac.cn}

\author{Kai Shu}
\affiliation{%
  \institution{Illinois Institute of Technology}
  \city{Chicago}
  \state{Illinois}
  \country{USA}
}
\email{kshu@iit.edu}


\begin{abstract}

Emotion plays an important role in detecting fake news online. When leveraging emotional signals, the existing methods focus on exploiting the emotions of news contents that conveyed by the publishers (i.e., \textit{publisher emotion}). However, fake news often evokes high-arousal or activating emotions of people, so the emotions of news comments aroused in the crowd (i.e., \textit{social emotion}) should not be ignored. Furthermore, it remains to be explored whether there exists a relationship between \textit{publisher emotion} and \textit{social emotion} (i.e., \textit{dual emotion}), and how the \textit{dual emotion} appears in fake news. In this paper, we verify that \textit{dual emotion} is distinctive between fake and real news and propose \DEE~to represent \textit{dual emotion} and the relationship between them for fake news detection. Further, we exhibit that our proposed features can be easily plugged into existing fake news detectors as an enhancement. Extensive experiments on three real-world datasets (one in English and the others in Chinese) show that our proposed feature set: 1) outperforms the state-of-the-art task-related emotional features; 2) can be well compatible with existing fake news detectors and effectively improve the performance of detecting fake news.\footnote{Please kindly note that the examples in this paper contain offensive and swear words.} \footnote{The code and datasets are released at https://github.com/RMSnow/WWW2021.}
\end{abstract}

\maketitle

\section{Introduction}
In recent years, fake news on social media has threatened not only cyberspace security, but also the real-world order in politics \cite{fisher2016pizzagate}, economy \cite{elboghdady2013market}, society \cite{Bangladesh-lynchings}, etc. 
The most recent example is the concomitant \textit{infodemic} during the COVID-19 pandemic across the world \cite{infodemic}. Thousands of news pieces with misleading content have been spreading through social media \cite{wiki-covid-19} and led to socio-economic disorder \cite{covid-shuanghuanglian} and weakened the effect of pandemic prevention \cite{covid-public-health}. 
To tackle this issue, researchers have been devoted to developing automatic methods to detect fake news (i.e., designing a classifier to judge a given news piece as real or fake) by leveraging signals from text \cite{castillo-www11, qazvinian2011rumor, perez2018automatic}, images \cite{jin2016novel, mvnn}, or social contexts \cite{shu2018understanding, shu-beyond-contents, li2019rumor, majing-data, ma2017detect, ma2018rumor, han-cikm}. \footnote{In this paper, we use \textit{news pieces} to refer to social media news posts. A news piece generally contains content and its attached comments.} 

\begin{figure}
  \begin{subfigure}{\linewidth}
    \centering
    \includegraphics[width=\linewidth]{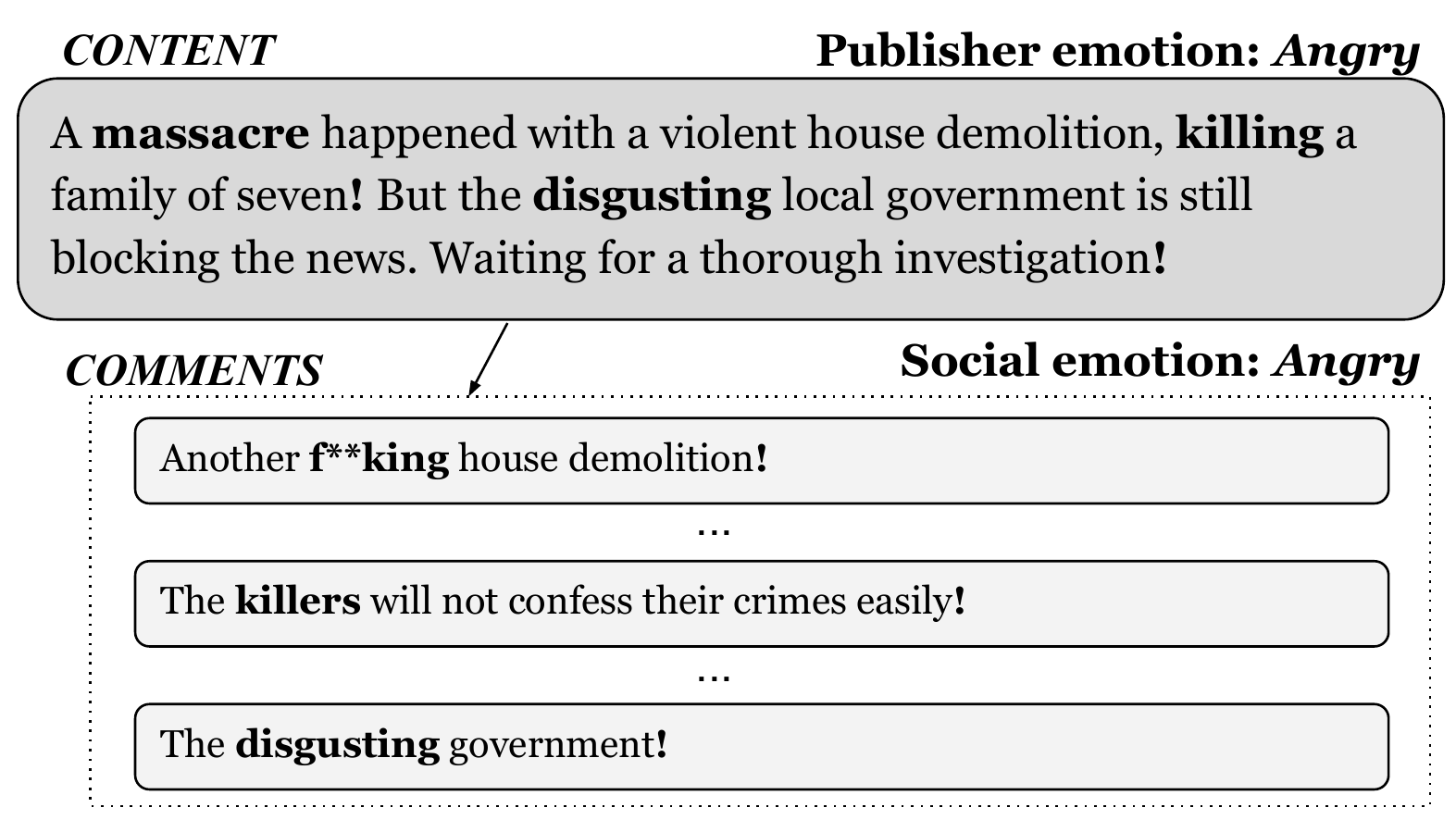}  
    \caption{Emotion resonance in a fake news piece: the \textit{publisher emotion} and \textit{social emotion} are both \textit{angry}.}
    \label{fig:intro-cases-1}
  \end{subfigure}
  \begin{subfigure}{\linewidth}
    \centering
    \includegraphics[width=\linewidth]{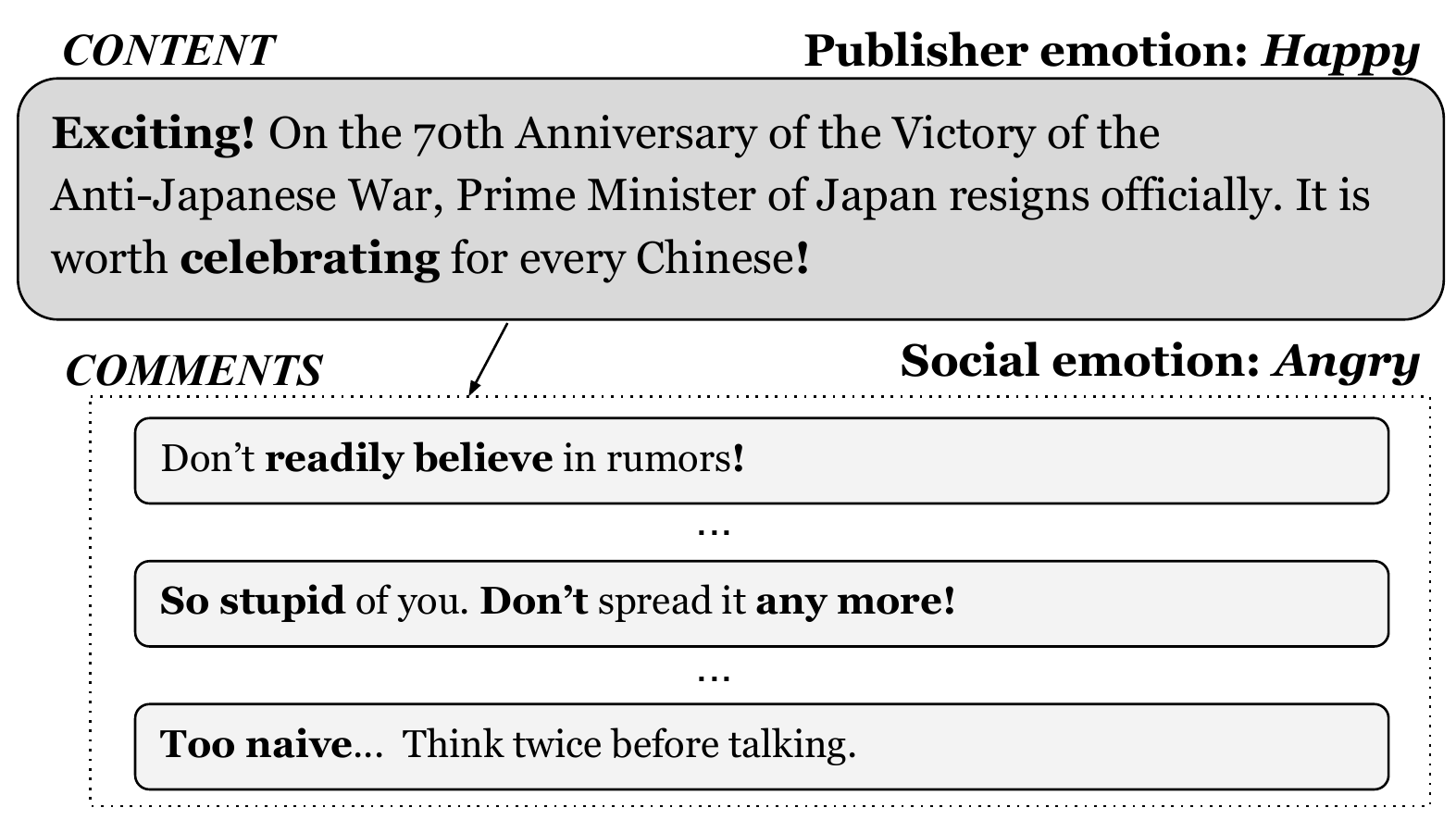}  
    \caption{Emotion dissonance in a fake news piece: the \textit{publisher emotion} is \textit{happy} while the \textit{social emotion} is \textit{angry}.}
    \label{fig:intro-cases-2}
  \end{subfigure}
\caption{Two fake news pieces on Chinese microblog platform Weibo, with different \textit{Dual Emotion}. The texts are translated from Chinese to English manually.}
\label{fig:intro-cases}
\end{figure}

\begin{figure*}[t]
  \centering
  \includegraphics[width=\textwidth]{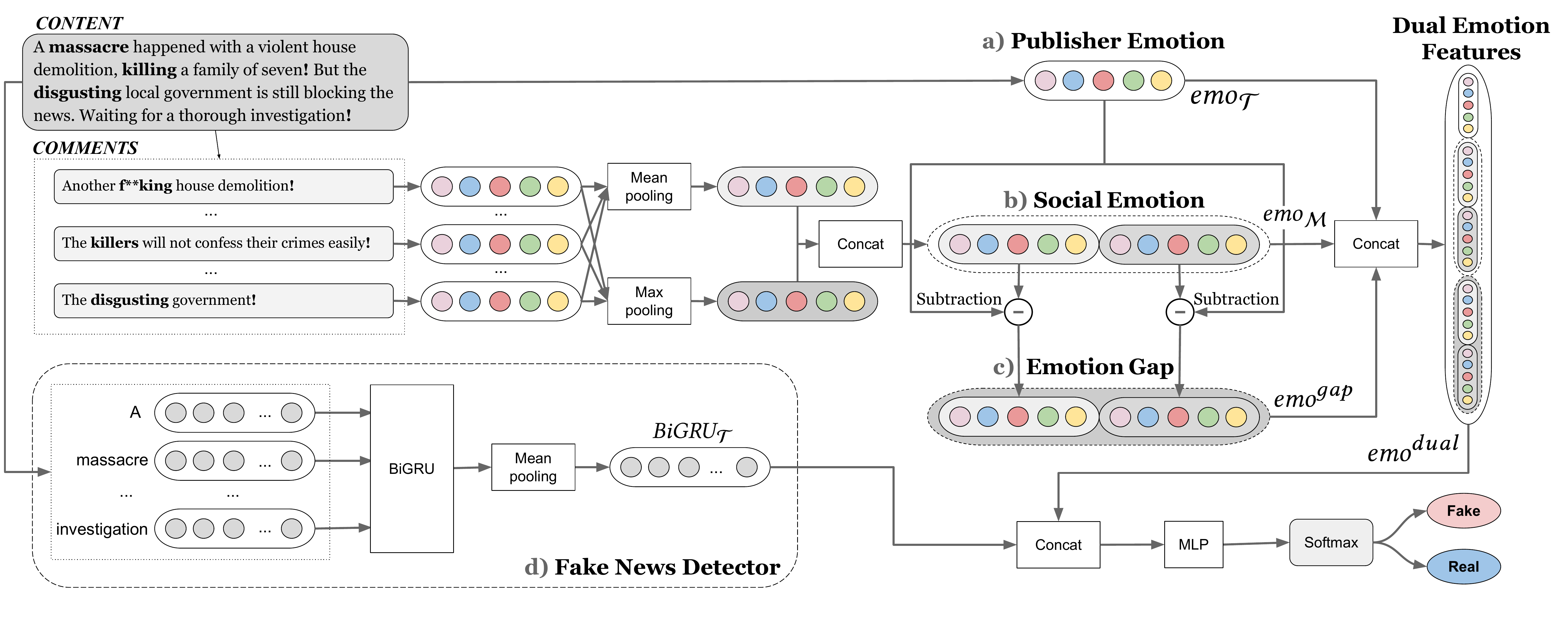}
  \caption{An overall framework of using \DEE~ for fake news detection. \DEE~ consist of three components: a) \textit{Publisher Emotion} extracted from the content; b) \textit{Social Emotion} extracted from the comments; c) \textit{Emotion Gap} representing the similarity and difference between publisher emotion and social emotion. \DEE~ are concatenated with the features from d) Fake News Detector (here, BiGRU as an example) for the final prediction of veracity.}
  \label{fig:dee}
\end{figure*}

In existing text-based works \cite{castillo-www11, emotion-icassp, emotion-sigir}, the role of sentimental or emotional signals has been considered for fake news detection. \citet{emotion-icassp} point out that there exists a relationship between news veracity and the sentiments of the posted text, and append a sentimental feature (the ratio of the number of negative and positive words) to help text-only fake news detectors. Instead of appending a sole feature, \citet{emotion-sigir} extract richer emotional features from the news contents based on emotional lexicons for fake news detection. To the best of our knowledge, most existing works leverage the emotional signals of fake news content conveyed by the publishers but rarely focus on the emotions of fake news comments aroused in the crowd. However, for spreading in the crowd virally, fake news often evokes high-arousal or activating emotions of the crowd \cite{fake-news-rosnow}. Therefore, in addition to emotions of news contents, it is necessary to explore whether emotions of news comments and the relationship between the two emotions are helpful for fake news detection.

To describe the two emotions clearly, we define them respectively as 1) \textbf{\textit{publisher emotion}}: the emotions conveyed by publishers of the news pieces; and 2) \textbf{\textit{social emotion}}: the emotions aroused in the crowd facing to the news pieces. And we adopt \textbf{\textit{dual emotion}} as a general term of these two emotions. 
For a news piece, \textit{dual emotion} has two appearances: emotion resonances (i.e., the \textit{publisher emotion} is same or similar to the \textit{social emotion}) and emotion dissonances (i.e., the \textit{publisher emotion} is different from the \textit{social emotion}). 
We analyze the data and find that the two appearances have a statistically significant distinction between fake and real news (see details in Section~\ref{sec:analysis}). 
For example, as to the emotion resonance, there are more fake news pieces whose \textit{dual emotion} are both \textit{angry} than real news, while as to the emotion dissonances, more fake news pieces whose \textit{publisher emotion} is \textit{happy} while \textit{social emotion} is \textit{angry}.
Figure~\ref{fig:intro-cases} shows two representative examples selected from fake news pieces on Weibo \footnote{https://www.weibo.com}. 
In Figure~\ref{fig:intro-cases-1}, the fake news publisher conveys its rage with expressions like ``massacre'', ``killing'', ``disgusting''. As a result, the great indignation of the crowd is evoked, shown by ``f**king'', ``killers'', and ``disgusting''. 
In Figure~\ref{fig:intro-cases-2}, the fake news publisher expresses happiness with ``Exciting!'' and ``celebrating''. While the crowd considers it as a ridiculous news piece, and use ``readily believe'', ``So stupid'' and ``Too naive'' to express their disgust and contempt to the publisher.
The data observation statistical findings highlight that the relationship in \textit{dual emotion} can be indicative of the news veracity and should be considered when modeling.

To model the \textit{dual emotion} and emotion resonances and dissonances for fake news detection, we propose \DEE~ to represent \textit{publisher emotion}, \textit{social emotion} and the similarity and difference of the \textit{dual emotion} jointly. Besides, it is convenient to implement and plug the features into existing fake news detectors as an enhancement.

In this paper, our contributions are summarized as follows:

\begin{itemize}
    \item We propose and verify that the \textit{dual emotion} (i.e., \textit{publisher emotion} and \textit{social emotion}) signal is distinctive between fake and real news.
    \item We firstly propose the feature set, \DEE, to comprehensively represent dual emotion and the relationship between the two kinds of emotions, and exhibit how to plug it into the fake news detectors as a complement and enhancement.
    \item We conduct experiments on the real-world datasets, including a newly-constructed Chinese dataset. The results demonstrate that: 1) \DEE~ outperforms the existing emotional features for fake news detection. 2) It can be compatible with existing fake news detectors and effectively improve the performance of the detectors.
\end{itemize}
\section{Related Work}
Fake news detection is also known as false news detection, rumor detection, misinformation detection, etc. \cite{survey-false-news} and is closely connected to the field of information credibility evaluation. In the earliest study on information credibility evaluation, \citet{castillo-www11} manually extract content features, publisher features, topic features, and propagation features from news pieces. And the work finds that sentiment-based features like the fraction of sentimental words and exclamation marks are effective for evaluating information credibility. In recent years, researchers begin to utilize deep learning models such as GRU-based and CNN-based models for fake news detection \cite{majing-data, cami-ijcai}. Beyond news content, social contexts such as texts of comments and reposts \cite{majing-data, ma2017detect, ma2018rumor, han-cikm, csi-cikm}, viewpoints and stances of the crowd \cite{ours-aaai16, branchLSTM}, and user credibility\cite{shu-beyond-contents, li2019rumor} are emphasized as well.

There are also existing works focusing on discovering the distinctive emotional signals between fake and real news. \citet{emotion-icassp} verify that there exists a relationship between news veracity (real or fake) and the usage of sentimental words, and design an emotion feature (the ratio of the count of negative and positive words) to help detect fake news. Besides, \citet{emotion-sigir} extract emotion features based on emotional lexicons from news contents for fake news detection. However, these works only leverage the emotional signals of fake news contents but ignore the emotions of fake news comments and the relationship between the two emotions. Recently, \citet{fusion-ecai20} propose an adaptive fusion network for fake news detection, modeling emotion embeddings from the contents and the comments. However, this work focuses on adaptively fusing various features by advanced deep learning models, and do not explore the specific distinction of dual emotion signals between fake and real news. So far, the work that pays attention to mining dual emotion signals from publishers and crowds remains vacant.
\section{Modeling Dual Emotion for Fake News Detection} \label{sec:model}
To model dual emotion signals for fake news detection, we propose \DEE, which can leverage publisher emotion, social emotion, and the similarity and difference of the dual emotion.
Figure~\ref{fig:dee} exhibits the process of obtaining \DEE~and integrating them into an existing fake news detector as an enhancement to classify a given piece of news. In this section, we detail the feature extraction of publisher emotion and social emotion, and the modeling of emotion gap. ~Then, we describe the process to plug \DEE into the existing fake news detectors.

\subsection{Publisher Emotion} \label{sec:pe}
To comprehensively represent the \textit{Publisher Emotion}, we use a variety of features extracted from news contents, including the emotion category, emotional lexicon, emotional intensity, sentiment score, and other auxiliary features. In the five kinds of features, emotion category, emotional intensity and sentiment score provide the overall information and the other two provide word- and symbol-level information. 

Given the input sequence of the textual content with length $L$, $\mathcal{T} = [t_1, t_2, \ldots, t_i, \ldots, t_L]$, where $t_i$ is the $i^{th}$ word in the text, the goal is to extract emotion features $emo_\mathcal{T}$ from the text $\mathcal{T}$.

\subsubsection{Emotion Category}
We use public emotion classifiers (which will be introduced in Section~\ref{sec:analysis}) to get emotion category features. Usually, the output of an emotion classifier is the probabilities that the given text contains certain emotions.

Given the emotion classifier $f$ and the text $\mathcal{T}$, we assume the dimension of the output is $d_f$ and thus the prediction of the text is $f(\mathcal{T})$.
So we can obtain the emotion category features $emo_\mathcal{T}^{cate} = f(\mathcal{T})$, where $emo_\mathcal{T}^{cate} \in \mathds{R}^{d_f}$.

\subsubsection{Emotional Lexicon}\label{para:lexicon}
Usually, a piece of text conveys specific emotions by using several specific words (which are generally included in emotional lexicons). Thus, we next extract the features based on the emotional lexicon. The approach is dependent on the existing emotion dictionaries annotated by experts. In the emotion dictionary, we assume that there are $d_e$ kinds of emotions, denoted as $E = \{e_1, e_2, \ldots, e_{d_e}\}$. For the emotion $e \in E$, the dictionary provides a list of emotional words $\mathscr{E}_e = \{w_{e,1}, w_{e,2}, \ldots, w_{e,L_e}\}$, where $L_e$ is the length of the emotion lexicon of $e$ in the dictionary.

Given the text $\mathcal{T}$, we gradually aggregate the scores of each word and the whole text across all the emotions for rich representation. For one of the emotions $e$, we firstly calculate the word-level score $s(t_i, e)$, where $t_i$ is $i^{th}$ word in the text $\mathcal{T}$.  If the word $t_i$ is in the dictionary $\mathscr{E}_e$, we consider not only its occurrence frequency, but also its contextual words (specifically, degree words and negation words). 
For example, in the sentence ``I am not very joyful today'' (the length of the sentence is 6), ``joyful'' belongs to the emotion \textit{happy} and its occurrence frequency is 1/6. Assume that we only consider the left context and the window size is 2 (i.e., the context words are ``not'' and ``very''). When we set the negation value of ``not'' as -1 and the degree value of ``very'' as 2, the final $s(joyful, e_{happy}) = -1 * 2 * (1/6) = -1/3$. In practice, we use the existing emotion dictionary to match and calculate the values of negation and degree words. As described above, $s(t_i, e)$ is defined in Equation~\ref*{eq:lexicon-score1}:

\begin{equation}\label{eq:lexicon-score1}
s(t_i, e) = \frac{\mathds{1}_{\mathscr{E}_e}(t_i) * neg(t_i, w) * deg(t_i, w)}{L}
\end{equation}

\begin{equation}\label{eq:lexicon-score2}
 \mathds{1}_{\mathscr{E}_e}(t_i) = 
    \left\{
      \begin{aligned}
      1, & &if~t_i \in \mathscr{E}_e \\
      0, & &otherwise
      \end{aligned}
    \right.
\end{equation}

where $w$ is the window size of the left context. And $neg(t_j)$ (Equation~\ref*{eq:lexicon-score3}) and $deg(t_j)$ (Equation~\ref*{eq:lexicon-score4}) are respectively the negation value and degree value of $t_j$, which can be looked up according to the emotion dictionary.

\begin{equation}\label{eq:lexicon-score3}
neg(t_i, w) = \prod_{j=i-w}^{i-1} neg(t_j)
\end{equation}

\begin{equation}\label{eq:lexicon-score4}
deg(t_i, w) = \prod_{j=i-w}^{i-1} deg(t_j)
\end{equation}

We then calculate text-level score on the specific emotion $e$, denoted as $s(\mathcal{T}, e)$, by summing the scores of each word in the text, as Equation~\ref*{eq:lexicon1} shows: 
\begin{equation} \label{eq:lexicon1}
s(\mathcal{T}, e) = \sum_{i=1}^L s(t_i, e), \quad \forall e \in E
\end{equation}

Finally, the emotional lexicon features $emo_\mathcal{T}^{lex}$ are obtained by concatenating all the scores of the $d_e$ emotions (Equation~\ref*{eq:lexicon2}), where $\oplus$ is the concatenation operator, and $emo_\mathcal{T}^{lex} \in \mathds{R}^{d_e}$.

\begin{equation} \label{eq:lexicon2}
emo_\mathcal{T}^{lex} = s(\mathcal{T}, e_1) \oplus s(\mathcal{T}, e_2) \oplus \cdot \cdot \cdot \oplus s(\mathcal{T}, e_{d_e}) \\
\end{equation}

\subsubsection{Emotional Intensity}
As for emotional lexicons, we also consider the emotional intensity of the lexicons. For example, when expressing the emotion $happy$, the word ``ecstatic'' owns a higher intensity than ``joyful''. The extracting process is similar to that of the emotional lexicon features, except for that we here include the intensity scores. Given the emotions $E$, the emotional word list $\mathscr{E}_e$ for every emotion $e$, and the text $\mathcal{T}$, we first calculate the intensity-aware text-level scores $s^{\prime}(\mathcal{T}, e)$ by summing the intensity-weighted word-level scores, as shown in Equation~\ref*{eq:intensity1}:
\begin{equation} \label{eq:intensity1}
    s^{\prime}(\mathcal{T}, e) = \sum_{i=1}^L s^{\prime}(t_i, e) = \sum_{i=1}^L int(t_i) * s(t_i, e), \quad \forall e \in E
\end{equation}
where $int(t_i)$ denotes the intensity score of the word $t_i$. If $t_i$ is in the dictionary, $int(t_i)$ can be calculated according to the emotion dictionary, otherwise $int(t_i) = 0$.

The emotional intensity features $emo_\mathcal{T}^{int}$ can be obtained  by concatenating all the intensity scores of $d_e$ kinds of emotions, as shown in Equation~\ref*{eq:intensity2}: 

\begin{equation} \label{eq:intensity2}
    emo_\mathcal{T}^{int} = s^{\prime}(\mathcal{T}, e_1) \oplus s^{\prime}(\mathcal{T}, e_2) \oplus \cdot \cdot \cdot \oplus s^{\prime}(\mathcal{T}, e_{d_e}) \\
\end{equation}
where $emo_\mathcal{T}^{int} \in \mathds{R}^{d_e}$.

\subsubsection{Sentiment Score}
In addition to the emotion-level features described above, we also consider the coarse-grained sentiment score of the text. Usually, the sentiment score is a positive or negative value, which represents the degree of the positive or negative polarity of the whole text. And it can be calculated by using sentiment dictionaries or public toolkits. Assuming that the dimension of the sentiment score is $d_s$ (usually, $d_s = 1$), we can get the sentiment score feature $emo_\mathcal{T}^{senti} \in \mathds{R}^{d_s}$.

\subsubsection{Other Auxiliary Features}\label{sec:aux}
Considering that the above features do not explicitly exploit the information beyond emotion dictionaries, we introduce a set of auxiliary features to capture the emotional signals behind the non-word elements, including emoticons, punctuations, and uppercase letters (only for English). Also, we add the frequency of sentimental words and personal pronouns to enhance the awareness of the users' word usages. Take emoticons as an example. The emoticons are universal for emotional expression across the world, such as ``: )'' for $happy$, ``: ('' for $sad$. Besides, punctuations like ``!'' and ``?'' can also convey people's moods and emotions. Table~\ref{tab:aux} summarizes the auxiliary features used in the \DEE. Assume that there are $d_a$ features, and we can extract the other auxiliary features $emo_\mathcal{T}^{aux} \in \mathds{R}^{d_a}$.

\begin{table}[h]
  \centering
  \resizebox{\columnwidth}{!}{%
  \begin{tabular}{c|l}
  \hline
  \textbf{Type} & \multicolumn{1}{c}{\textbf{Features}} \\ \hline
  \multirow{5}{*}{Emoticons} & The frequency of happy emoticons \\
   & The frequency of angry emoticons \\
   & The frequency of surprised emoticons \\
   & The frequency of sad emoticons \\
   & The frequency of neutral emoticons \\ \hline
  \multirow{3}{*}{Punctuations} & The frequency of exclamation mark \\
   & The frequency of question mark \\
   & The frequency of ellipsis mark \\ \hline
  \multirow{4}{*}{Sentimental Words} & The frequency of positive sentimental words \\
   & The frequency of negative sentimental words \\
   & The frequency of degree words \\
   & The frequency of negation words \\ \hline
  \multirow{3}{*}{Personal Pronoun} & The frequency of pronoun first \\
   & The frequency of pronoun second \\
   & The frequency of pronoun third \\ \hline
  \begin{tabular}[c]{@{}c@{}}Others\\ (For English corpus)\end{tabular} & The frequency of uppercase letters \\ \hline
\end{tabular}%
}
\caption{Auxiliary Feature List}
\label{tab:aux}
\end{table}

To get the \textit{Publisher Emotion} of the text $\mathcal{T}$ from the content, we concatenate all five kinds of features described above and obtain $emo_\mathcal{T}$, as shown in Equation~\ref*{eq:pe}:

\begin{equation} \label{eq:pe}
  emo_\mathcal{T} = emo_\mathcal{T}^{cate} \oplus emo_\mathcal{T}^{lex} \oplus emo_\mathcal{T}^{int} \oplus emo_\mathcal{T}^{senti} \oplus emo_\mathcal{T}^{aux}
\end{equation}
where $emo_\mathcal{T} \in \mathds{R}^{d}$ (i.e., $d = d_f + 2 d_e + d_s + d_a$).

\subsection{Social Emotion}
We first extract \textit{Social Emotion} from the comments of a news piece and then aggregate them as the whole representation. The comments of a news piece are denoted as $\mathcal{M} = [\mathcal{M}_1, \mathcal{M}_2, \ldots, \mathcal{M}_i, \ldots, \\\mathcal{M}_{L_\mathcal{M}}]$, where $\mathcal{M}_i$ is the $i^{th}$ comment of the news piece, and $L_\mathcal{M}$ is the length of comment list. As for $\mathcal{M}_i$, we can calculate its emotion vector $emo_{\mathcal{M}_i}$ by Equation~\ref*{eq:pe}, where $emo_{\mathcal{M}_i} \in \mathds{R}^d$. Then we stack the transposed emotion vector (row vector) of every comment to obtain the whole emotion vector of comments $\widehat{emo_{\mathcal{M}}}$, as shown in Equation~\ref{eq:se1}:
\begin{equation}\label{eq:se1}
	\widehat{emo_{\mathcal{M}}} = emo_{\mathcal{M}_1}^\mathsf{T} \oplus emo_{\mathcal{M}_2}^\mathsf{T} \oplus \cdot \cdot \cdot \oplus emo_{\mathcal{M}_{L_\mathcal{M}}}^\mathsf{T}
\end{equation}
where $\widehat{emo_{\mathcal{M}}}  \in \mathds{R}^{L_\mathcal{M} \times d}$.

After getting $\widehat{emo_{\mathcal{M}}}$, we consider two aggregators to generate the \textit{Social Emotion} of the whole comment list: 1) Mean pooling for representing the average emotional signals (Equation~\ref{eq:se2}); and 2) max pooling for capturing the extreme emotional signals (Equation~\ref{eq:se3}).

\begin{equation}\label{eq:se2}
emo_{\mathcal{M}}^{mean} = mean(\widehat{emo_{\mathcal{M}}})
\end{equation}

\begin{equation}\label{eq:se3}
emo_{\mathcal{M}}^{max} = max(\widehat{emo_{\mathcal{M}}}) 
\end{equation}
where $emo_{\mathcal{M}}^{mean},  emo_{\mathcal{M}}^{max} \in \mathds{R}^{d}$.

 Finally, we concatenate them as the \textit{Social Emotion}:
 \begin{equation}\label{eq:se4}
 emo_{\mathcal{M}} = emo_{\mathcal{M}}^{mean} \oplus emo_{\mathcal{M}}^{max}
 \end{equation}
where $emo_{\mathcal{M}} \in \mathds{R}^{2d}$.

\subsection{Emotion Gap}
To model the resonances and dissonances of dual emotion, we propose \textit{Emotion Gap} (denoted as $emo^{gap}$). It is designed as the subtraction between \textit{Publisher Emotion} and \textit{Social Emotion}. As shown in Equation~\ref*{eq:gap}, $emo^{gap}$ is concatenated by the difference of $emo_\mathcal{T}$ and $emo_\mathcal{M}^{mean}$ and the difference of $emo_\mathcal{T}$ and $emo_\mathcal{M}^{max}$:
\begin{equation} \label{eq:gap}
  emo^{gap} = (emo_\mathcal{T} - emo_\mathcal{M}^{mean}) \oplus (emo_\mathcal{T} - emo_\mathcal{M}^{max})
\end{equation}
where $emo^{gap} \in \mathds{R}^{2d}$. By this means, it can measure the differences (i.e., dissonances) between the dual emotion. For emotions resonances, the values in the \textit{Emotion Gap} vector are tiny (nearly zero).

\subsection{Dual Emotion Features}
Finally, \DEE~are concatenated by the \textit{Publisher Emotion}, the \textit{Social Emotion} and the \textit{Emotion Gap}. In Equation~\ref*{eq:dual} we obtain the \DEE, where $emo^{dual} \in \mathds{R}^{5d}$.

\begin{equation} \label{eq:dual}
  emo^{dual} = emo_\mathcal{T} \oplus emo_\mathcal{M} \oplus emo^{gap}
\end{equation}

After getting \DEE, we can concatenate it with representations that extracted by the fake news detectors, which is exemplified by Figure~\ref{fig:dee}. Assuming that the fake news detector is BiGRU and the output feature vector is denoted as  $BiGRU_{\mathcal{T}}$, the concatenated vector $[BiGRU_{\mathcal{T}}, emo^{dual}]$ is fed into a multi-layer perceptron (MLP) layer and a softmax layer for the final prediction of news veracity $\hat{y}$, as shown in Equation \ref{eq:final}:
\begin{equation} \label{eq:final}
	\hat{y} = {\rm Softmax}\big({\rm MLP}([BiGRU_{\mathcal{T}}, emo^{dual}])\big)
\end{equation}

\section{Experiments and Evaluation}

In this section, we conduct experiments to compare our proposed \DEE~ and other baseline features and explore their roles in improving the performance of fake news detection. Specifically, we mainly answer the following evaluation questions:

\begin{itemize}
	\item \textbf{EQ1:} Are \DEE~ more effective than baseline features when used alone for fake news detection? How effective are the different types of features in \DEE?
	\item \textbf{EQ2:} Can \DEE~ help improve the performance of text-based fake news detectors?
	\item \textbf{EQ3:} How robust do the fake news detection models with \DEE~ in real-world scenarios?
	\item \textbf{EQ4:} How effective are the components of \DEE, including the publisher emotion, social emotion, and emotion gap?
\end{itemize}

\subsection{Dataset} \label{sec:dataset}

\begin{table*}[t]
  \centering
  \small
  \begin{tabular}{c|c|rr|rr|rr}
      \hline
      \multirow{2}{*}{\textbf{}} & \multirow{2}{*}{\textbf{Veracity}} & \multicolumn{2}{c|}{\textbf{RumourEval-19}} & \multicolumn{2}{c|}{\textbf{Weibo-16}} & \multicolumn{2}{c}{\textbf{Weibo-20}} \\ \cline{3-8} 
       &  & \multicolumn{1}{c}{\textbf{\#pcs}} & \multicolumn{1}{c|}{\textbf{\#com}} & \multicolumn{1}{c}{\textbf{\#pcs}} & \multicolumn{1}{c|}{\textbf{\# com}} & \multicolumn{1}{c}{\textbf{\#pcs}} & \multicolumn{1}{c}{\textbf{\#com}} \\ \hline
      \multirow{4}{*}{\textbf{Training}} & Fake & 79 & 1,135 & 801 & 649,673 & 1,896 & 749,141 \\
       & Real & 144 & 1,905 & 1,410 & 482,226 & 1,920 & 516,795 \\
       & Unverified & 104 & 1,838 & - & - & - & - \\ \cline{2-8} 
       & \textbf{Total} & 327 & 4,878 & 2,211 & 1,131,899 & 3,816 & 1,265,936 \\ \hline
      \multirow{4}{*}{\textbf{Validating}} & Fake & 19 & 824 & 268 & 222,149 & 632 & 137,941 \\
       & Real & 10 & 404 & 470 & 146,948 & 640 & 185,087 \\
       & Unverified & 9 & 212 & - & - & - & - \\ \cline{2-8} 
       & \textbf{Total} & 38 & 1,440 & 738 & 369,097 & 1,272 & 323,028 \\ \hline
      \multirow{4}{*}{\textbf{Testing}} & Fake & 40 & 689 & 286 & 193,740 & 633 & 245,216 \\
       & Real & 31 & 805 & 471 & 179,942 & 641 & 149,260 \\
       & Unverified & 10 & 181 & - & - & - & - \\ \cline{2-8} 
       & \textbf{Total} & 81 & 1,675 & 757 & 373,682 & 1,274 & 394,476 \\ \hline
      \multirow{4}{*}{\textbf{Total}} & Fake & 138 & 2,648 & 1,355 & 1,065,562 & 3,161 & 1,132,298 \\
       & Real & 185 & 3,114 & 2,351 & 809,116 & 3,201 & 851,142 \\
       & Unverified & 123 & 2,231 & - & - & - & - \\ \cline{2-8} 
       & \textbf{Total} & 446 & 7,993 & 3,706 & 1,874,678 & 6,362 & 1,983,440 \\ \hline
      \end{tabular}%
\caption{Statistics of the three datasets. \#pcs: number of news pieces; \#com: number of comments.}
\label{tab:dataset}
\end{table*}

Although the emotions are believed universal, albeit affected by culture \cite{eckman}, how emotions are expressed and perceived varies across different socio-cultural backgrounds \cite{culture-emotion}. Thus, we conduct experiments on three real-world datasets in two languages (meanwhile, two countries with different cultures), one in English (\textit{RumourEval-19}) and two in Chinese (\textit{Weibo-16} and \textit{Weibo-20}). The statistics of these datasets are shown in Table~\ref*{tab:dataset}.

\subsubsection{RumourEval-19}
The dataset \textit{RumourEval-19} is constructed for determining the veracity of the rumors on Twitter and Reddit. It is released in an academic evaluation\footnote{SemEval-2019 Task 7: http://alt.qcri.org/semeval2019/index.php?id=tasks} \cite{RumourEval-data}. Each news piece is labeled as fake, real, or unverified. We keep the same dataset splits and evaluation criteria as what the organizers provide.

\subsubsection{Weibo-16}\label{sec:weibo-16}
The dataset \textit{Weibo-16} is firstly proposed in \cite{majing-data} and has been a benchmark dataset of fake news detection  in Chinese \cite{csi-cikm, cami-ijcai, han-cikm}. Each news piece is labeled as fake or real. It needs to be clarified that in the original dataset, the subset of fake news has many duplications. Concerned about the influence to learning and evaluation by duplications, we perform deduplication on the subset of fake news based on a clustering algorithm based on text similarity. As a result, the amount of clusters is only 59\% of the original amount of fake pieces. We suppose that the duplication may increase the risk of data leakage when splitting training and testing sets and make models tend to learn some event-specific features\cite{eann} (as they may repeat multiple times in the training process), which limits the generalizability of models. Therefore, we filtered out the highly similar fake news pieces and produce a deduplication version of \textit{Weibo-16} (Table~\ref*{tab:dataset}). We also clustered real news pieces but found no duplications in \textit{Weibo-16}. As an empirical supplement of our analysis, we conduct comparison experiments between the original and the deduplication version of \textit{Weibo-16}, and verified the necessity of deduplication (see details in Appendix A). In our experiments in the main text, the deduplicated \textit{Weibo-16} is divided into train / val. / test sets in the ratio of 3:1:1.

\subsubsection{Weibo-20}
As a benchmark Chinese dataset for fake news detection, \textit{Weibo-16} contains fake news pieces ranging from Dec 2010 to April 2014, and is not extended until now. Besides, the scale of \textit{Weibo-16} is smaller after deduplication (Section~\ref*{sec:weibo-16}). Therefore, we constructed the dataset \textit{Weibo-20} on the basis of \textit{Weibo-16}.

We keep the two-class setting (i.e., fake or real for each news pieces). For fake news, we retain the 1,355 fake news pieces of \textit{Weibo-16} and further collect news pieces judged as misinformation officially by Weibo Community Management Center\footnote{https://service.account.weibo.com/} (the same source of fake news of \textit{Weibo-16} \cite{majing-data}) ranging from April 2014 to Nov 2018. And we filter out the highly similar fake news pieces and guarantee there are no duplications. For real news, we retain the 2,351 real news pieces of \textit{Weibo-16} and gather 850 unique real news pieces in the same period as the fake news. The newly-collected real news pieces are real news verified by NewsVerify\footnote{https://www.newsverify.com/} which focuses on discovering and verifying suspicious news pieces on Weibo. Totally, \textit{Weibo-20} contains 3,161 fake news pieces and 3,201 real news pieces. As for dataset splits, we split train / val. / test sets in the ratio of 3:1:1.

\subsection{Preliminary Analysis of Dual Emotion Signals}
\label{sec:analysis}

\begin{figure*}[t]
  \centering
  \begin{subfigure}{.27\textwidth}
    \centering
    \includegraphics[width=\linewidth]{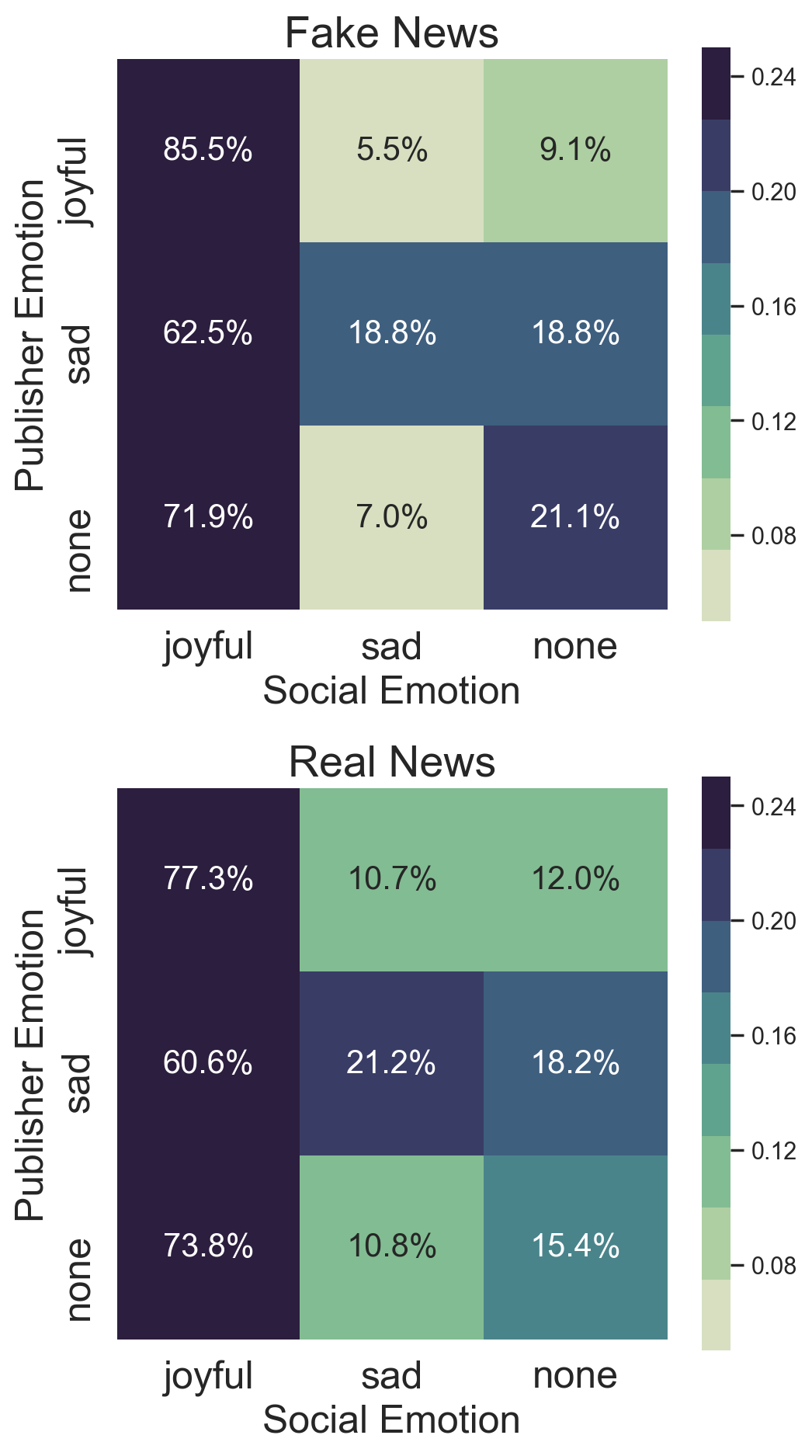}  
    \caption{On \textit{RumourEval-19}}
    \label{fig:dual-emotion-rumoureval}
  \end{subfigure}
  \hspace{.04\textwidth}
  \begin{subfigure}{.27\textwidth}
    \centering
    \includegraphics[width=\linewidth]{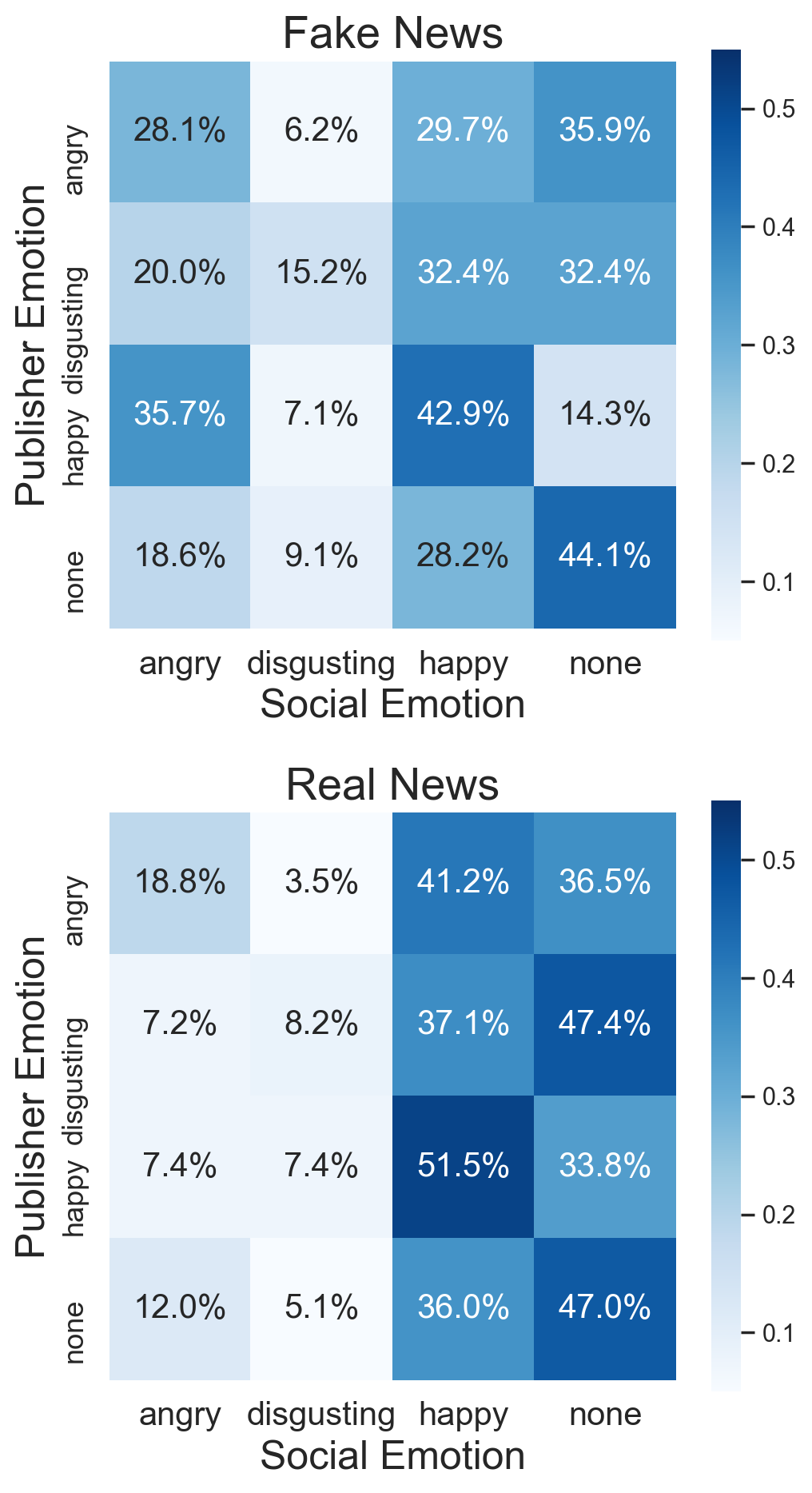}  
    \caption{On \textit{Weibo-16}}
    \label{fig:dual-emotion-majing-weibo}
  \end{subfigure}
  \hspace{.04\textwidth}
  \begin{subfigure}{.27\textwidth}
      \centering
      \includegraphics[width=\linewidth]{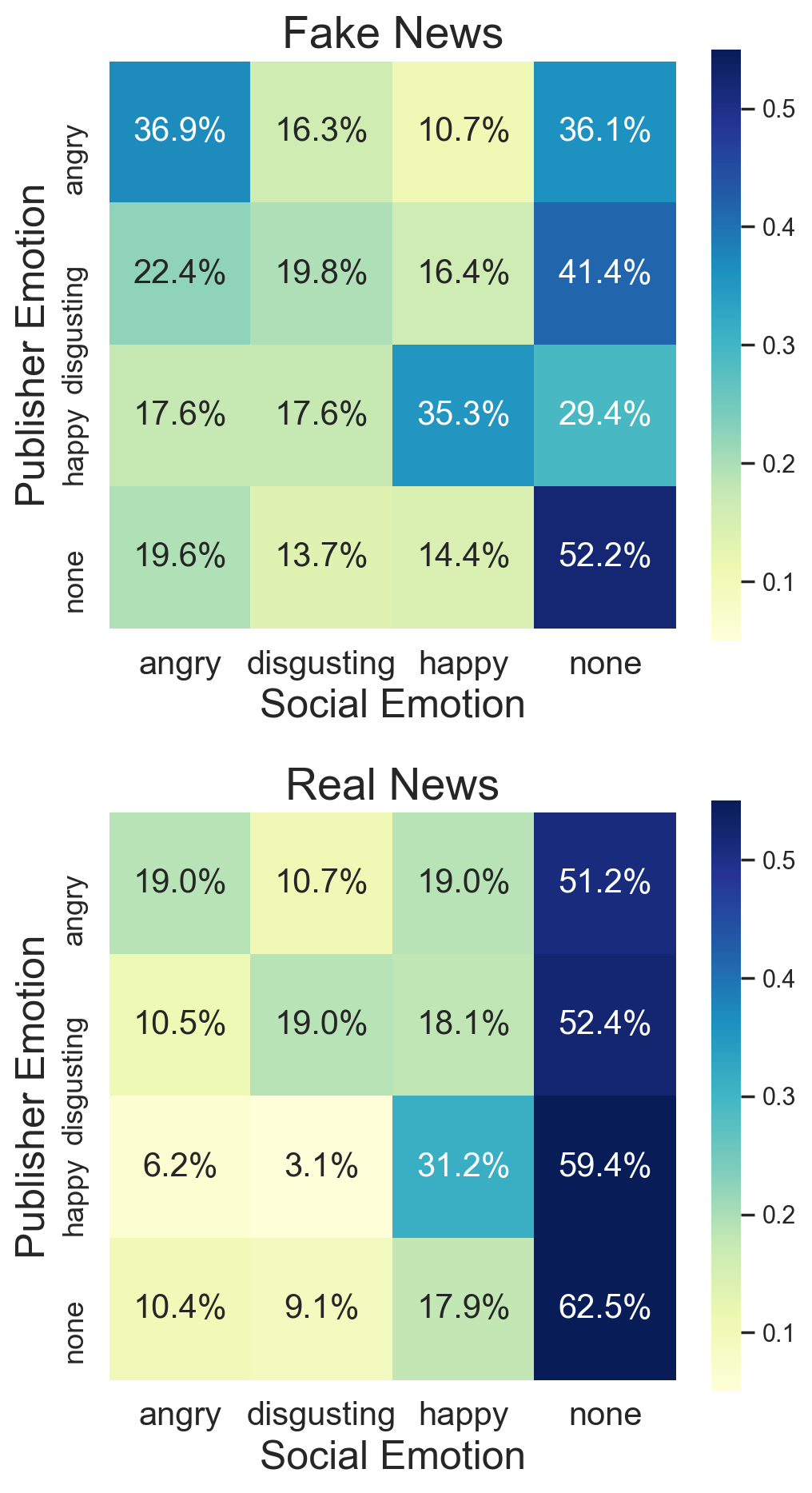}  
      \caption{On \textit{Weibo-20}}
      \label{fig:dual-emotion-weibo}
    \end{subfigure}
\caption{The distribution of \textit{Dual Emotion Category} on the three datasets. In fake news, there are distinct emotion resonances and emotion dissonances from real news.}
\label{fig:dual-emotion-category}
\end{figure*}


To check whether it is statistically dependent or not between dual emotion signals and the veracity of news pieces, we construct two categorical variables to do a chi-squared statistical significance test. The one is \textit{News Veracity}, whose value is \textit{Fake} or \textit{Real}. The other is \textit{Dual Emotion Category}, whose value is combined publisher emotion category and social emotion category, such as \textit{publisher emotion is none and social emotion is angry}. To calculate the value of \textit{Dual Emotion Category}, we use the open-source emotion classification model released by NVIDIA\footnote{https://github.com/NVIDIA/sentiment-discovery} \cite{nvidia-emotion} for \textit{RumourEval-19}, and use \textit{Emotion Detection Service} on Baidu AI platform\footnote{https://ai.baidu.com/tech/nlp/emotion\_detection} for the two Chinese datasets.
In the chi-squared statistical significance test, we firstly assume that the dual emotion signals are independent of the veracity of news pieces (i.e., the null hypothesis). 
Then we check whether the chi-squared statistic is over the critical value or not. 
Specifically, on the dataset \textit{RumourEval-19}, the chi-squared statistic is 50.570, over the critical value of 48.602 for the probability of 95\%, which means we can reject the null hypothesis. 
Similarly, on the dataset \textit{Weibo-16}, the chi-squared statistic is 209.14, which is much more than the critical value of 50.892 for the probability of 99\%. And on the dataset \textit{Weibo 20}, the chi-squared statistic is 239.963, which is much more than the critical value of 46.963 for the probability of 99\%. 
In conclusion, we can reject the null hypothesis on all three datasets, which indicates that dual emotion signals are statistically dependent on news veracity.

We visualize the variable \textit{Dual Emotion Category} further. On \textit{RumourEval 19}, we select three emotion categories to visualize, \textit{joyful}, \textit{sad} and \textit{none} (over 98\% of news pieces covered). And on Chinese datasets, we select four emotion categories, \textit{angry}, \textit{disgusting}, \textit{happy} and \textit{none} (over 97\% of news pieces covered). We utilize the heatmap to exhibit the distribution of \textit{Dual Emotion Category} in Figure~\ref*{fig:dual-emotion-category}. In the heatmap, each cell represents the percentage of news pieces whose \textit{Dual Emotion Category} is the specific value. And we normalize the percentages for each row (i.e., each publisher emotion). For example, in the top sub-figure of Figure~\ref*{fig:dual-emotion-rumoureval}, the upper-left cell indicates that among fake news pieces whose publisher emotion is \textit{joyful}, the percentage of pieces whose social emotion is also \textit{joyful} is 85.5\%.

In Figure~\ref{fig:dual-emotion-category}, we can see there are distinct emotion resonances and emotion dissonances in fake news from real news. For example, in Figure~\ref*{fig:dual-emotion-rumoureval}, the percentage of dual emotion categories that are both \textit{joyful} in fake news is 8.2\% higher than that of real news. And the percentage of emotion dissonance with \textit{sad} publisher emotion and \textit{joyful} social emotion in fake news is 1.9\% higher than real news. Evidence is stronger on the two Chinese datasets. Specifically, as for emotion resonances, there are more news pieces whose dual emotion categories are both \textit{angry} and are both \textit{disgusting} in fake news than real news.
As for emotion dissonances, there are more news pieces emotion dissonances with are \textit{happy}/\textit{none} publisher emotion but \textit{angry}/\textit{disgusting} social emotion in fake news.

It needs to be recognized that the specific emotion resonances or dissonances may vary from English to Chinese datasets, since the expression styles of people using different languages may be also different. However, our analysis shows that on each dataset itself, no matter what its \textit{dominant language} is, the fake news owns distinct emotion resonances and dissonances from real news, which can be helpful for distinguishing the fake and real news.

\subsection{Experimental Setup}

\subsubsection{Emotion Resources}
\label{sec:resources}
For emotion classifiers, as described in Section~\ref{sec:analysis}, we adopt the pretrained models of NVIDIA for English and Baidu AI for Chinese. To ensure the robustness of the two models, per language we randomly sampled 100 instances and had their emotion categories manually and independently labeled by three annotators, resulting the accuracy of 87\% for NVIDIA model and 83\% for Baidu model. Therefore, the two classifiers are considered reliable for extracting emotions for fake news detection. As for other emotion resources, for English corpus, we adopt NRC Emotion lexicon\cite{nrc-lexicon} and NRC Emotion Intensity lexicon\cite{nrc-intensity} to extract emotion lexicon and emotion intensity features, respectively. And we use the Vader package of NLTK\cite{nltk} to calculate sentiment scores. For Chinese corpus, we adopt the Affective Lexicon Ontology\cite{dalianligong-dict} to extract emotion lexicon and emotion intensity features. And we utilize the dictionary HowNet\cite{hownet} to calculate sentiment scores. As for auxiliary features in Table~\ref*{tab:aux}, for emoticons, we utilize the \textit{List of emoticons} of Wikipedia\cite{wiki-emoticons} and divide emoticons into five emotions: \textit{happy}, \textit{angry}, \textit{surprised}, \textit{sad} and \textit{neutral}. For sentimental words and degree words, we use the bilingual sentiment dictionary in HowNet\cite{hownet}. For negation words, we compile the words list from Wikipedia, Oxford Dictionary, and Cambridge Dictionary.\footnote{The negation word lists are released together with our code and datasets.}

\subsubsection{Fake News Detectors and Baselines}
In the experiments, we select two baseline emotion features to evaluate the effectiveness of our \DEE. These features are implemented with the same emotion dictionaries as \DEE:

\begin{itemize}
    \item \textbf{Emoratio}: \citet{emotion-icassp} propose an emotion feature that can be extracted from the content text of news pieces, named \textit{emoratio}. It is calculated by the ratio of \textit{count of negative emotional words} and \textit{count of positive emotional words}. 
    \item \textbf{EmoCred}: \citet{emotion-sigir} utilize the emotional lexicon and intensity features of the content texts. These features are calculated based on the lexicons' occurrence frequency. 
\end{itemize}

For testing the ability of the emotional features to help the text-based fake news detectors (especially those that do not explicitly model the emotional signals), we select BiGRU (as Figure~\ref{fig:dee} shows), BERT, and other state-of-the-art fake news detectors as follows:

\begin{itemize}
    \item \textbf{BiGRU}: 
    Text-based models like GRU\cite{gru} and LSTM\cite{lstm} are proven effective for fake news detection in \cite{majing-data, callAtRumor-pakdd}. Here we use \textbf{BiGRU} to examine whether \DEE~can improve it or not. In practice, as for word embeddings, we use GloVe \cite{GloVe} for English and Chinese Word Vectors for Chinese \cite{chinese-w2v}. The max sequence length of $BiGRU_{\mathcal{T}}$ is 100, and the dimensionality of hidden state of $BiGRU_{\mathcal{T}}$ is 32.
    \item \textbf{BERT} \cite{BERT}: As a strong text classification model, \textbf{BERT} has been adopted to represent semantic signals when detecting fake news in \cite{fusion-ecai20}. In the experiments, we truncate the sequences to the maximum length of 512, and finetune the pretrained models\footnote{The pretrained models are downloaded from https://huggingface.co/models. We use \textit{bert-base-uncased} for English and \textit{bert-base-chinese} for Chinese.} for our task.
    \item \textbf{NileTMRG}~\cite{NileTMRG}: For \textit{RumourEval-19} dataset, we use the model implemented by the competition organizers\footnote{https://github.com/kochkinaelena/RumourEval2019}~\cite{RumourEval-data}, \textbf{NileTMRG}. The model is effective and outperforms other contestants' models of the leaderboard except for the champion.
The model is a linear SVM and uses text features, social features, and use comment stance features. In practice, we keep all the hyperparameters of the original model.
    \item \textbf{HSA-BLSTM}~\cite{han-cikm}: For the two Chinese datasets, we implement the \textbf{HSA-BLSTM}, which is widely used as a baseline on \textit{Weibo-16} dataset. The authors propose a hierarchical attention neural network and utilize not only the contents of news pieces but also the comments. 
    In experiments, we keep all the hyperparameters as those in the original model.
\end{itemize}

\subsubsection{Model Parameters}
The dimensionalities of sub features in \DEE, i.e., $d_f$, $d_e$, $d_s$ and $d_a$, are determined by the language-specific emotion resources. The value of $d_f$, as the output of pretrained emotion classifiers, is 16 for English and 8 for Chinese. The value of $d_e$ is the size of emotion kinds of the English or Chinese emotion dictionaries, which is 8 or 21, respectively. For $d_s$, sentiment scores of English texts, produced by the Vader package of NLTK, correspond to four dimensions (positive, negative, neutral and compound), while sentiment scores of Chinese texts are calculated by HowNet, which have one dimension only. The value of $d_a$ is the number of the heuristic features in Table~\ref{tab:aux}, which is 16 for English and 15 for Chinese. The full dimension $d$ is computed as Equation~\ref{eq:pe}, which is 52 for English and 66 for Chinese. The window size is 2, which was determined by grid search that maximizes the performance on the validation set. As for the amount of comments, we set $L_\mathcal{M} = 100$, which means that only the earliest 100 comments (or less) of every news piece are considered. In Equation~\ref*{eq:final}, the output dimensionality of $\rm MLP$ is 32.



\subsubsection{Evaluation Metrics} 
On \textit{RumourEval-19}, we adopt the official evaluation metrics, macro F1 score and RMSE (root mean squared error)~\cite{RumourEval-data}. Considering the imbalance of the dataset, we also consider the F1 scores of fake, real, and unverified news. On the two Weibo datasets, we use accuracy and macro F1 score as the evaluation metrics, the same as \cite{han-cikm}. We also the F1 scores of fake and real news. The other experiments use the macro F1 score.

\subsection{Results}

\subsubsection{Effectiveness of Dual Emotion Features}

To answer \textbf{EQ1} under the circumstance that the confounding factor of fake news detectors is excluded, we utilize emotion features \textbf{alone} to detect fake news. We adopt a simple five-layer MLP and feed only emotion features into it. Table~\ref{tab:results-only-emotion} displays the results on the three datasets.

\begin{table}[h]
    \centering
    \begin{tabular}{p{0.16\linewidth}|p{0.39\linewidth}|p{0.09\linewidth}|p{0.09\linewidth}|p{0.09\linewidth}}
        \hline
        \textbf{Source} & \multicolumn{1}{l|}{\textbf{Emotion Features}} & \textbf{R-19} & \textbf{W-16} & \textbf{W-20} \\ \hline
        \multirow{3}{*}{Content} & Emoratio & 0.185 & 0.553 & 0.524 \\
         & EmoCred & 0.253 & 0.564 & 0.542 \\
         & \textbf{Publisher Emotion} & 0.290 & 0.571 & 0.573 \\ \hline
        Comments & \textbf{Social Emotion} & 0.296 & 0.692 & 0.754 \\ \hline
        \multirow{2}{*}{\begin{tabular}[c]{@{}c@{}}Content, \\ Comments\end{tabular}} & \textbf{Emotion Gap} & 0.332 & 0.716 & 0.746 \\
         & \textbf{Dual Emotion Features} & \textbf{0.337} & \textbf{0.728} & \textbf{0.759} \\ \hline
        \end{tabular}%
    \caption{Macro F1 scores when only using emotion features on the MLP model. R-19: RumourEval-19, W-16: Weibo-16, W-20: Weibo-20.}
    \label{tab:results-only-emotion}
\end{table}

\begin{table}[h]
    \centering
    \begin{tabular}{l|c|c|c}
        \hline
        \multicolumn{1}{c|}{\textbf{Removed type}} & \textbf{R-19} & \textbf{W-16} & \textbf{W-20} \\ \hline
        Emotion Category & 0.193 & 0.679 & 0.686 \\
        Emotion Lexicon & 0.239 & 0.715 & 0.745 \\
        Emotional Intensity & 0.216 & 0.725 & 0.750 \\
        Sentiment Score & 0.245 & 0.723 & 0.743 \\
        Other Auxiliary Features & 0.307 & 0.653 & 0.722 \\ \hline
        \end{tabular}%
    \caption{Macro F1 scores of \DEE~when removing one specific type of emotion features on the MLP model. R-19: RumourEval-19, W-16: Weibo-16, W-20: Weibo-20.}
    \label{tab:results-emotion-types}
\end{table}

\begin{table*}[t]
    \centering
        \begin{tabular}{l|c|c|ccc}
            \hline
            \multicolumn{1}{c|}{\multirow{2}{*}{\textbf{Models}}} & \multirow{2}{*}{\textbf{Macro F1 score}} & \multirow{2}{*}{\textbf{RMSE}} & \multicolumn{3}{c}{\textbf{F1 score}} \\ \cline{4-6} 
            \multicolumn{1}{c|}{} &  &  & \textbf{Fake News} & \textbf{Real News} & \textbf{Unverified News} \\ \hline
            BiGRU & 0.269 & 0.804 & 0.500 & 0.222 & 0.083 \\
             + Emoratio & 0.275 & 0.823 & 0.463 & 0.160 & \textbf{0.200} \\
             + EmoCred & 0.311 & 0.797 & 0.456 & 0.295 & 0.182 \\
             + \textbf{Dual Emotion Features}  & \textbf{0.340} & \textbf{0.752} & \textbf{0.580} & \textbf{0.337} & 0.104 \\ \hline
            BERT & 0.272 & 0.808 & 0.533 & 0.105 & 0.176 \\
             + Emoratio & 0.271 & 0.857 & 0.406 & 0.240 & 0.167 \\
             + EmoCred & 0.308 & 0.833 & 0.367 & \textbf{0.367} & 0.189 \\
             + \textbf{Dual Emotion Features}  & \textbf{0.346} & \textbf{0.778} & \textbf{0.557} & 0.244 & \textbf{0.238} \\ \hline
            NileTMRG & 0.309 & 0.770 & 0.557 & 0.245 & 0.125 \\
             + Emoratio & 0.331 & \textbf{0.754} & \textbf{0.571} & 0.280 & \textbf{0.143} \\
             + EmoCred & 0.307 & 0.786 & 0.296 & 0.500 & 0.125 \\
             + \textbf{Dual Emotion Features}  & \textbf{0.342} & \textbf{0.754} & 0.565 & \textbf{0.565} & 0.100 \\ \hline
        \end{tabular}%
    \caption{Results on \textit{RumourEval-19}.}
    \label{tab:results-rumoureval}
\end{table*}

\begin{table*}[t!]
    \centering
    \begin{tabular}{l|c|c|cc|cccc}
        \hline
        \multicolumn{1}{c|}{\multirow{3}{*}{\textbf{Models}}} & \multicolumn{4}{c|}{\textbf{Weibo-16}} & \multicolumn{4}{c}{\textbf{Weibo-20}} \\ \cline{2-9} 
        \multicolumn{1}{c|}{} & \multirow{2}{*}{\textbf{Macro F1 score}} & \multirow{2}{*}{\textbf{Accuracy}} & \multicolumn{2}{c|}{\textbf{F1 score}} & \multicolumn{1}{c|}{\multirow{2}{*}{\textbf{Macro F1 score}}} & \multicolumn{1}{c|}{\multirow{2}{*}{\textbf{Accuracy}}} & \multicolumn{2}{c}{\textbf{F1 score}} \\ \cline{4-5} \cline{8-9} 
        \multicolumn{1}{c|}{} &  &  & \textbf{Fake} & \textbf{Real} & \multicolumn{1}{c|}{} & \multicolumn{1}{c|}{} & \textbf{Fake} & \textbf{Real} \\ \hline
        BiGRU & 0.807 & 0.822 & 0.754 & 0.860 & \multicolumn{1}{c|}{0.839} & \multicolumn{1}{c|}{0.839} & 0.839 & 0.839 \\
         + Emoratio & 0.794 & 0.810 & 0.738 & 0.851 & \multicolumn{1}{c|}{0.850} & \multicolumn{1}{c|}{0.850} & 0.854 & 0.846 \\
         + EmoCred & 0.766 & 0.778 & 0.711 & 0.820 & \multicolumn{1}{c|}{0.829} & \multicolumn{1}{c|}{0.829} & 0.836 & 0.821 \\
         + \textbf{Dual Emotion Features} & \textbf{0.826} & \textbf{0.838} & \textbf{0.781} & \textbf{0.871} & \multicolumn{1}{c|}{\textbf{0.855}} & \multicolumn{1}{c|}{\textbf{0.855}} & \textbf{0.857} & \textbf{0.852} \\ \hline
        BERT & 0.824 & 0.845 & 0.762 & 0.886 & \multicolumn{1}{c|}{0.900} & \multicolumn{1}{c|}{0.900} & 0.900 & 0.900 \\
         + Emoratio & 0.837 & 0.857 & 0.780 & 0.894 & \multicolumn{1}{c|}{0.901} & \multicolumn{1}{c|}{0.901} & 0.900 & 0.902 \\
         + EmoCred & 0.849 & 0.867 & 0.797 & \textbf{0.901} & \multicolumn{1}{c|}{0.902} & \multicolumn{1}{c|}{0.902} & 0.901 & 0.903 \\
         + \textbf{Dual Emotion Features} & \textbf{0.867} & \textbf{0.873} & \textbf{0.837} & 0.896 & \multicolumn{1}{c|}{\textbf{0.915}} & \multicolumn{1}{c|}{\textbf{0.915}} & \textbf{0.913} & \textbf{0.918} \\ \hline
        HSA-BLSTM & 0.849 & 0.855 & 0.819 & 0.879 & \multicolumn{1}{c|}{0.913} & \multicolumn{1}{c|}{0.913} & 0.912 & 0.914 \\
         + Emoratio & 0.863 & 0.872 & 0.829 & 0.898 & \multicolumn{1}{c|}{0.920} & \multicolumn{1}{c|}{0.920} & 0.920 & 0.920 \\
         + EmoCred & 0.854 & 0.861 & 0.822 & 0.886 & \multicolumn{1}{c|}{0.903} & \multicolumn{1}{c|}{0.903} & 0.902 & 0.905 \\
         + \textbf{Dual Emotion Features} & \textbf{0.908} & \textbf{0.913} & \textbf{0.885} & \textbf{0.930} & \multicolumn{1}{c|}{\textbf{0.932}} & \multicolumn{1}{c|}{\textbf{0.932}} & \textbf{0.932} & \textbf{0.933} \\ \hline
        \end{tabular}%
\caption{Results on \textit{Weibo-16} and \textit{Weibo-20}.}
\label{tab:results-weibo}
\end{table*}
\vspace{-5pt}

In Table~\ref*{tab:results-only-emotion}, among the three emotion features that source from Content, \textit{Publisher Emotion} is more effective than \textit{EmoCred} and \textit{Emoratio}, especially on \textit{RumourEval}. It reveals the effectiveness of \DEE~in modeling emotional signals. What's more, we can see the more improvements of \textit{Social Emotion} and \textit{Emotion Gap}, which are first proposed to help detect fake news in this paper. Specifically, on \textit{RumourEval-19}, using \textit{Emotion Gap} owns 4.2\% increase than \textit{Publisher Emotion}. And on the two Chinese datasets, using \textit{Social Emotion} or \textit{Emotion Gap} can both improve the macro F1 score of more than 10\%. Moreover, using \DEE~can further obtain enhancements on the three datasets. Especially on \textit{RumourEval-19}, only using \DEE~for fake news detection owns a high macro F1 score of 0.337. And only using \textit{Emotion Gap} is also effective, which is 0.332 for the macro F1 score. It is worth mentioning that such two emotion features even outperform the state-of-the-art model \textbf{NileTMRG} (0.309 for macro F1 score, shown in Table~\ref{tab:results-rumoureval}). That indicates the necessity of dual emotion signals and the importance of mining dual emotion and the relationship between them for fake news detection. Additionally, it needs to be clarified that comparing the three datasets to each other, the performances in \textit{RumourEval-19} are rather worse than the two Chinese datasets. The reasons are discussed in \cite{RumourEval-data,eventAI}, that the amount of news pieces is small and there is a relatively low inter-annotator agreement for the dataset. 

In Section~\ref*{sec:pe}, we adopt five types of emotion features when modeling emotional signals (Emotion Category, Emotion Lexicon, Emotional Intensity, Sentiment Score, and Other Auxiliary Features). To verify the effect of every type of emotion features, we remove one specific type of features from \DEE~every time, to observe the performance changes. As Table~\ref*{tab:results-emotion-types} shows, the macro F1 scores of \DEE~ all decrease regardless of the removed type of emotion features. Thus, it reveals the necessity of using five types of emotion features jointly.

\subsubsection{Performance Evaluation within Fake News Detectors}

To answer \textbf{EQ2}, we exhibit the results of adding \DEE~into the existing fake news detectors on the three datasets. 

Table~\ref{tab:results-rumoureval} exhibits the results on \textit{RumourEval-19} dataset. Overall, after using \DEE, the three fake news detectors are both improved a lot. Specifically, on the text-based detectors, \textbf{BiGRU} and \textbf{BERT}, the use of \DEE~both improves the performance more than \textit{EmoCred} and \textit{Emoratio}. Especially, putting \DEE~ into \textbf{BERT} owns 0.346 for macro F1 score, far more than the other two emotion features. On the state-of-the-art model \textbf{NileTMRG}, using \textit{Emoratio} and \DEE~both improves the macro F1 score further. And the improvement of \DEE~ is 3.3\%, which is 1.1\% higher than \textit{Emoratio}.

The experimental results on the two Weibo datasets are displayed in Table~\ref*{tab:results-weibo}. Overall, we can see that our proposed \DEE~outperforms \textit{Emoratio} and \textit{EmoCred} on any models in both datasets. Specifically, on \textbf{BiGRU} and \textbf{BERT}, the improvements in macro F1 score of \DEE~are at least 1.5\% higher on the two datasets. However, when using \textit{Emoratio} or \textit{EmoCred} on \textbf{BiGRU}, sometimes the metrics even decrease. It reveals that \textit{Emoratio} and \textit{EmoCred} are more likely to be overfitted, since both of them focus on the contents alone but ignore the comments. And learning dual emotion jointly can avoid this situation to some extent. On the state-of-the-art model \textbf{HSA-BLSTM}, after using \DEE~as an enhancement, all the metrics are improved further in both datasets. Especially in \textit{Weibo-16}, the accuracy and macro F1 score both own about 6\% improvement, far more than \textit{Emoratio} and \textit{EmoCred}.

\subsubsection{Evaluation Under Real-World Scenario Simulation}\label{sec:temporal-splits}

In the fields of fake news detection, when splitting datasets, most works just \textbf{shuffle} the datasets and split them into train / val. / test sets \cite{majing-data, csi-cikm, cami-ijcai, han-cikm}, including the datasets splits in Table~\ref*{tab:dataset}. The kind of data split can somehow prove the effectiveness of proposed methods, but also has a shortcoming: In the real-world scenarios, when a check-worthy news piece emerges, we only own the data previously-emerging to train the detector, which cannot be guaranteed when adopting the above data split. To answer \textbf{EQ3}, we simulate a real-world scenario by additionally performing a temporal data split, which means that instances in the train / val. / test sets are arranged in chronological order, to evaluate the ability of models to detect \textit{future} news pieces.

\begin{table}[h]
    \centering
    \begin{tabular}{l|p{0.14\linewidth}|p{0.1\linewidth}|cc}
    \hline
    \multicolumn{1}{c|}{\multirow{2}{*}{\textbf{Models}}} & \multirow{2}{*}{\textbf{Macro F1}} & \multirow{2}{*}{\textbf{Acc.}} & \multicolumn{2}{c}{\textbf{F1 score}} \\ \cline{4-5} 
    \multicolumn{1}{c|}{} &  &  & \multicolumn{1}{c|}{\textbf{Fake}} & \textbf{Real} \\ \hline
    BiGRU & 0.680 & 0.681 & \multicolumn{1}{c|}{0.694} & 0.666 \\
     + Emoratio & 0.628 & 0.632 & \multicolumn{1}{c|}{0.665} & 0.592 \\
     + EmoCred & 0.659 & 0.666 & \multicolumn{1}{c|}{0.709} & 0.609 \\
     + \textbf{Dual Emotion Features} & \textbf{0.701} & \textbf{0.702} & \multicolumn{1}{c|}{\textbf{0.714}} & \textbf{0.689} \\ \hline
    BERT & 0.722 & 0.728 & \multicolumn{1}{c|}{0.762} & 0.682 \\
     + Emoratio & 0.719 & 0.724 & \multicolumn{1}{c|}{0.757} & 0.681 \\
     + EmoCred & 0.725 & 0.728 & \multicolumn{1}{c|}{0.752} & \textbf{0.699} \\
     + \textbf{Dual Emotion Features} & \textbf{0.734} & \textbf{0.734} & \multicolumn{1}{c|}{\textbf{0.773}} & 0.692 \\ \hline
    HSA-BLSTM & 0.776 & 0.778 & \multicolumn{1}{c|}{0.796} & 0.686 \\
     + Emoratio & 0.771 & 0.774 & \multicolumn{1}{c|}{0.796} & 0.663 \\
     + EmoCred & 0.777 & 0.781 & \multicolumn{1}{c|}{0.806} & 0.646 \\
     + \textbf{Dual Emotion Features} & \textbf{0.805} & \textbf{0.808} & \multicolumn{1}{c|}{\textbf{0.827}} & \textbf{0.694} \\ \hline
    \end{tabular}%
    \caption{Results on \textit{Weibo-20} (temporal data split). Acc. is short for Accuracy.}
    \label{tab:generalization}
\end{table}

In this section, we adopt the dataset \textit{Weibo-20} and select the most recent 20\% news pieces of them as the testing set. Among the remaining 80\% news pieces, we next select the most recent 25\% of them for validation and let the others be the training set. The results on temporally split \textit{Weibo-20} are displayed in Table~\ref*{tab:generalization}. Compared with Table~\ref*{tab:dataset}, we can see that in Table~\ref*{tab:generalization} all the performances decrease a lot. It indicates that the temporal data-split strategy creates a more challenging scenario, because the topics and writing styles of newly arrived instances are likely to change over time. Such a scenario can somehow expose the drawback of existing techniques and it requires a model of higher generalizability to cope with novel instances.

Under this hard setting, the models with our proposed \DEE~ still outperform those with \textit{Emoratio} and \textit{EmoCred}. Sometimes the introduction of \textit{Emoratio} or \textit{EmoCred} even leads to a performance decrease. In contrast, using \DEE~still enhances both models and increases all the metrics, which reveals the effectiveness and generalization ability of \DEE~to some extent.

 
\subsubsection{Ablation Study}

To answer \textbf{EQ4}, we further conduct ablation experiments on \textit{RumourEval-19}, \textit{Weibo-16}, \textit{Weibo-20} and \textit{Weibo-20 (temporally)} (splitting datasets temporally, described in Section~\ref*{sec:temporal-splits}). The results are displayed in Table~\ref*{tab:ablation}.

\begin{figure*}[h]
    \centering
    \makebox[0pt][c]{\parbox{\textwidth}{%
        \begin{minipage}[t]{0.32\hsize}\centering
            \includegraphics[width=\linewidth]{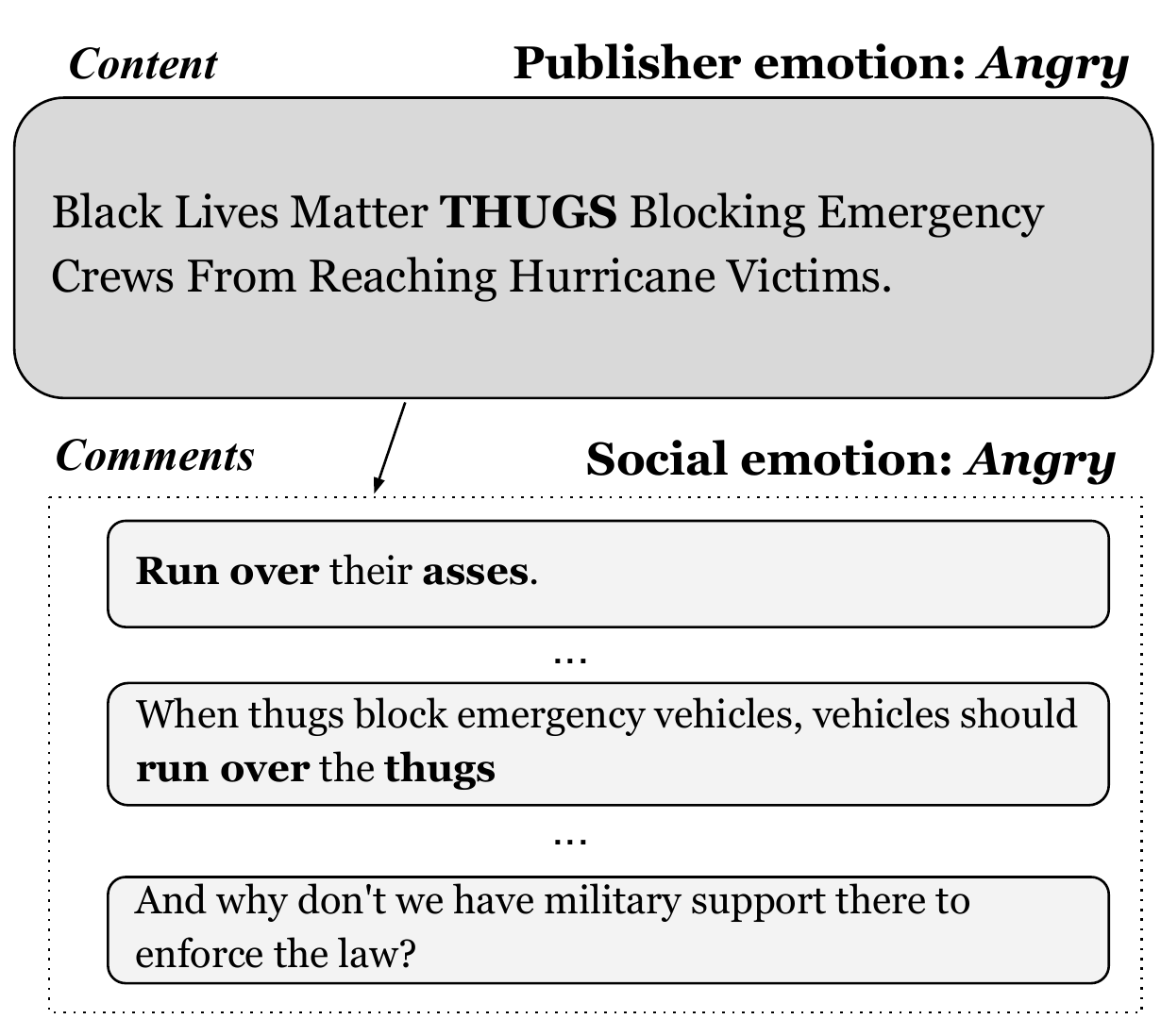}
        \end{minipage}
        \hfill
        \begin{minipage}[t]{0.32\hsize}\centering
            \includegraphics[width=\linewidth]{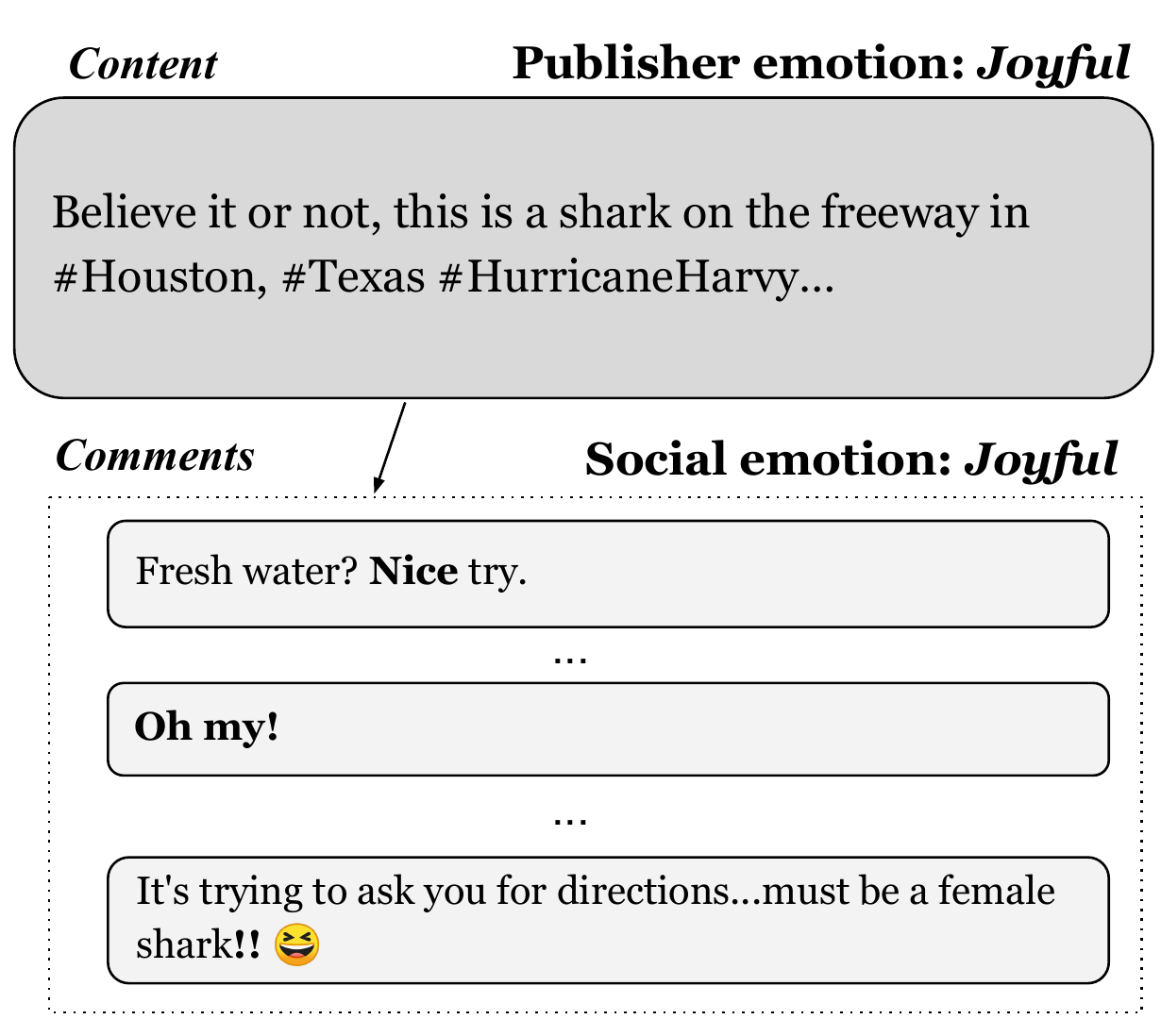}
        \end{minipage}
        \hfill
        \begin{minipage}[t]{0.32\hsize}\centering
            \includegraphics[width=\linewidth]{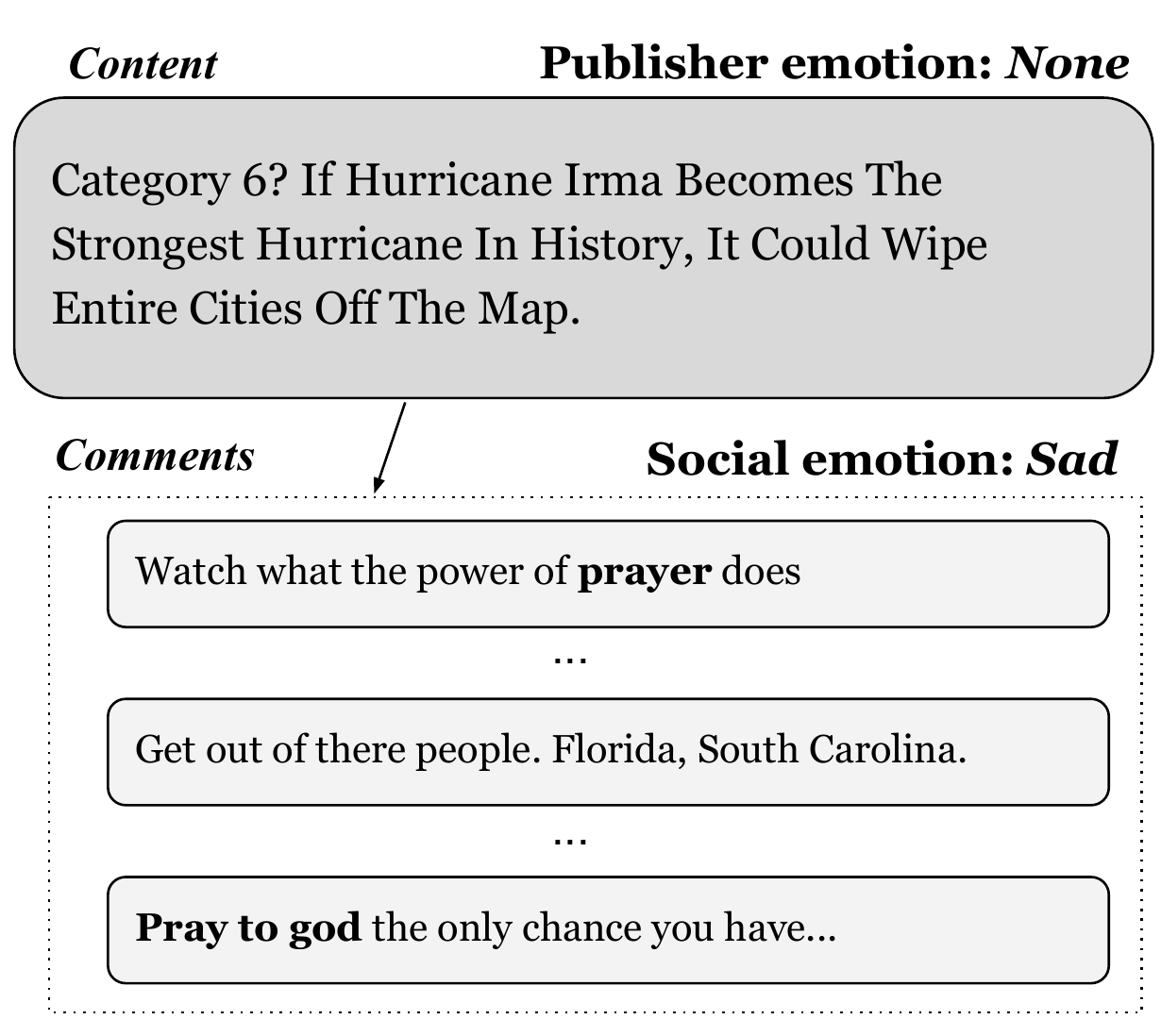}
        \end{minipage}
        \newline
        \begin{minipage}[t]{0.32\hsize}\centering
            \resizebox{\textwidth}{!}{%
            \begin{tabular}{lccc}
                & Fake & Real & Unverified \\
               BiGRU & 0.33 & \textbf{0.61} & 0.06 \\
               BiGRU + Emoratio & 0.35 & \textbf{0.57} & 0.08 \\
               BiGRU + EmoCred & 0.27 & \textbf{0.64} & 0.09 \\
               BiGRU + \textbf{Dual Emotion Features} & \textbf{0.65} & 0.21 & 0.14
               \end{tabular}%
            }
        \end{minipage}
        \hfill
        \begin{minipage}[t]{0.32\hsize}\centering
            \resizebox{\textwidth}{!}{%
            \begin{tabular}{lccc}
                & Fake & Real & Unverified \\
               BiGRU & 0.31 & \textbf{0.50} & 0.19 \\
               BiGRU + Emoratio & 0.36 & \textbf{0.56} & 0.08 \\
               BiGRU + EmoCred & 0.40 & \textbf{0.54} & 0.06 \\
               BiGRU + \textbf{Dual Emotion Features} & \textbf{0.65} & 0.22 & 0.13
               \end{tabular}%
            }
        \end{minipage}
        \hfill
        \begin{minipage}[t]{0.32\hsize}\centering
            \resizebox{\textwidth}{!}{%
            \begin{tabular}{lccc}
                & Fake & Real & Unverified \\
               BiGRU & 0.47 & \textbf{0.52} & 0.01 \\
               BiGRU + Emoratio & 0.40 & \textbf{0.59} & 0.01 \\
               BiGRU + EmoCred & 0.28 & \textbf{0.58} & 0.14 \\
               BiGRU + \textbf{Dual Emotion Features} & \textbf{0.63} & 0.17 & 0.20
               \end{tabular}%
            }
        \end{minipage}
    }}
\caption{Three fake news pieces on \textit{RumourEval-19}, which are missed by original BiGRU but detected after using \DEE. The prediction results of the four models are shown at the bottom, where the numbers represent confidence scores (a float value from 0 to 1). The scores that identify prediction labels are shown in bold.}
\label{fig:prediction-cases}
\end{figure*}

\begin{table}[h]
    \centering
    \begin{tabular}{p{0.11\linewidth}|p{0.28\linewidth}|c|c|c|c}
        \hline
        \multicolumn{2}{c|}{\textbf{Models}} & \textbf{R-19} & \textbf{W-16} & \textbf{W-20} & \textbf{W-20(t)} \\ \hline
        \multirow{4}{*}{BiGRU+} & Publisher Emotion & 0.310 & 0.809 & 0.842 & 0.681 \\
         & Social Emotion & 0.322 & 0.818 & 0.847 & 0.693 \\
         & Emotion Gap & 0.336 & 0.811 & 0.849 & 0.693 \\ \cline{2-6} 
         & \begin{tabular}{@{}l@{}}Dual Emotion \\ Features\end{tabular} & \textbf{0.340} & \textbf{0.826} & \textbf{0.855} & \textbf{0.701} \\ \hline
        \multirow{4}{*}{BERT+} & Publisher Emotion & 0.312 & 0.850 & 0.889 & 0.705 \\
         & Social Emotion & 0.339 & 0.856 & 0.911 & 0.730 \\
         & Emotion Gap & 0.338 & 0.858 & 0.906 & 0.731 \\ \cline{2-6} 
         & \begin{tabular}{@{}l@{}}Dual Emotion \\ Features\end{tabular} & \textbf{0.346} & \textbf{0.867} & \textbf{0.915} & \textbf{0.734} \\ \hline
        \multirow{4}{*}{\begin{tabular}[c]{@{}l@{}}Nile\\ TMRG+\end{tabular}} & Publisher Emotion & 0.311 & - & - & - \\
         & Social Emotion & 0.325 & - & - & - \\
         & Emotion Gap & 0.337 & - & - & - \\ \cline{2-6} 
         & \begin{tabular}{@{}l@{}}Dual Emotion \\ Features\end{tabular} & \textbf{0.342} & - & - & - \\ \hline
        \multirow{4}{*}{\begin{tabular}[c]{@{}l@{}}HSA-\\ BLSTM+\end{tabular}} & Publisher Emotion & - & 0.876 & 0.915 & 0.779 \\
         & Social Emotion & - & 0.892 & 0.922 & 0.792 \\
         & Emotion Gap & - & 0.901 & 0.926 & 0.800 \\ \cline{2-6} 
         & \begin{tabular}{@{}l@{}}Dual Emotion \\ Features\end{tabular} & - & \textbf{0.908} & \textbf{0.932} & \textbf{0.805} \\ \hline
        \end{tabular}%
    \caption{Ablation study of the three components of \DEE. The evaluation metric is macro F1 scores. R-19: RumourEval-19, W-16: Weibo-16, W-20: Weibo-20, and W-20(t): temporally split Weibo-20.}
    \label{tab:ablation}
\end{table}

In Table~\ref*{tab:ablation}, we can see that among the four datasets, adding \DEE~into the fake news detectors all obtain the highest macro F1 scores. Besides, compared with the original fake news detectors (Table~\ref*{tab:results-rumoureval} and Table~\ref*{tab:results-weibo}), using any component of \DEE~all enhances the performances of them. During the three components of \DEE, it exhibits that adopting \textit{Social Emotion} or \textit{Emotion Gap} improves the macro F1 scores more than \textit{Publisher Emotion} on any models on all the datasets. So it concludes that \textit{Social Emotion} and \textit{Emotion Gap} matter more when detecting fake news.

\subsection{Case Study}

We provide a qualitative analysis of \DEE~in some cases. Take the detector \textbf{BiGRU} on \textit{RumourEval-19} as an example, we select three fake news pieces that missed by the original \textbf{BiGRU} but detected after using \DEE~as an enhancement (Figure~\ref{fig:prediction-cases}). In the figure, there are rich dual emotion signals in every case, such as emotion resonances of \textit{angry} in the left case, of \textit{joyful} in the middle case, and emotion dissonances with \textit{none} publisher emotion and \textit{sad} social emotion in the right case. However, it exhibits using \textit{Emoratio} or \textit{EmoCred} do not help \textbf{BiGRU} detect rightly for the three cases. It reveals that mining dual emotion additionally sometimes is a remedy for the incompetence of only using semantics for detecting fake news.

\section{Conclusion and future work}
In this paper, we bring a new concept of \textit{dual emotion}, i.e., the publisher emotion and social emotion, into fake news research. We uncover the relationship between dual emotion signals (especially, the emotion gap) and the news veracity. Based on the data observation and analysis, we further propose a feature set, \DEE, to expose the distinctive emotional signals for detecting fake news. Further, we exhibit that our proposed features can be easily plugged into existing fake news detectors as an enhancement.
The extensive experiments conducted on three real-world datasets (including a newly-constructed Chinese dataset) have demonstrated that our proposed feature set outperforms the existing emotional features in fake news detection and essentially improves the performance of existing text-based methods. 
In future work, we plan to leverage multi-modal information (e.g., emotion in visual contents) to capture the emotions more precisely and use more sophisticated models for dual emotion representation.

\section*{Acknowledgments}
We thank Chuan Guo, Peng Qi, Yuting Yang for their insightful comments. This work is funded by National Natural Science Foundation of China (No. 61672523), and the Fundamental Research Funds for the Central Universities and the Research Funds of Renmin University of China (No. 18XNLG19). Kai Shu is supported by the John S. and James L. Knight Foundation through a grant to the Institute for Data, Democracy \& Politics at The George Washington University.

\bibliographystyle{ACM-Reference-Format}
\bibliography{sample-base-simple}

\section*{Appendix A. The reasons why the dataset \textit{Weibo-16} needs to be deduplicated}\label{sec:append_a}

In Section~\ref*{sec:weibo-16}, we mention that the original version of \textit{Weibo-16} contains many duplications of fake news pieces. Table~\ref*{tab:dataset-original-weibo16} shows the data statistics. Comparing to Table~\ref*{tab:dataset}, the number of fake news pieces decrease from 2,312 to 1,355 after deduplication. And there are no duplications in real news pieces.

\begin{table}[h]
    \centering
    \small
    \begin{tabular}{c|c|rr}
    \hline
    \textbf{} & \textbf{Veracity} & \multicolumn{1}{c}{\textbf{\#pcs}} & \multicolumn{1}{c}{\textbf{\#com}} \\ \hline
    \multirow{4}{*}{\textbf{Training}} & Fake & 1,386 & 789,841 \\
     & Real & 1,410 & 482,226 \\
     & Unverified & - & - \\ \cline{2-4} 
     & \textbf{Total} & 2,796 & 1,272,067 \\ \hline
    \multirow{4}{*}{\textbf{Validation}} & Fake & 463 & 255,833 \\
     & Real & 470 & 146,948 \\
     & Unverified & - & - \\ \cline{2-4} 
     & \textbf{Total} & 933 & 402,781 \\ \hline
    \multirow{4}{*}{\textbf{Testing}} & Fake & 463 & 224,795 \\
     & Real & 471 & 179,942 \\
     & Unverified & - & - \\ \cline{2-4} 
     & \textbf{Total} & 934 & 404,737 \\ \hline
    \multirow{4}{*}{\textbf{Total}} & Fake & 2,312 & 1,270,469 \\
     & Real & 2,351 & 809,116 \\
     & Unverified & - & - \\ \cline{2-4} 
     & \textbf{Total} & 4,663 & 2,079,585 \\ \hline
    \end{tabular}%
    \caption{Statistics of the original version of \textit{Weibo-16}. \#pcs: number of news pieces; \#com: number of comments.}
    \label{tab:dataset-original-weibo16}
\end{table}
\vspace{-0.5cm}

To further research the impact of duplications data on the ability of models, we conduct comparison experiments on the original and deduplicated versions of \textit{Weibo-16} respectively. And the results are exhibited in Table~\ref*{tab:results-weibo16-deduplication}. Here we choose \textbf{BiGRU} and \textbf{HSA-BLSTM} as fake news detectors. Considering the class imbalance of the deduplicated version of the dataset, we train the models based on class weights on the deduplicated training set.

\begin{table}[h]
    \centering
    \begin{tabular}{p{0.18\linewidth}|c|c|cc}
        \hline
        \multirow{2}{*}{\textbf{Models}} & \multicolumn{2}{c|}{\textbf{Dataset Version}} & \multirow{2}{*}{\textbf{Macro F1}} & \multirow{2}{*}{\textbf{Acc.}} \\ \cline{2-3}
         & \textbf{Train \& Val} & \textbf{Test} &  &  \\ \hline
        \multirow{3}{*}{BiGRU} & original & original & 0.793 & 0.793 \\ \cline{2-5} 
         & \multirow{2}{*}{deduplicated} & original & \textbf{0.806} & \textbf{0.807} \\ \cline{3-5} 
         &  & deduplicated & 0.807 & 0.822 \\ \hline
        \multirow{3}{*}{HSA-BLSTM} & original & original & 0.854 & 0.854 \\ \cline{2-5} 
         & \multirow{2}{*}{deduplicated} & original & \textbf{0.873} & \textbf{0.873} \\ \cline{3-5} 
         &  & deduplicated & 0.849 & 0.855 \\ \hline
        \end{tabular}%
    \caption{Results of the comparison experiments on the original and deduplication versions of \textit{Weibo-16}. Acc. is short for Accuracy.}
    \label{tab:results-weibo16-deduplication}
\end{table}
\vspace{-0.5cm}

In Table~\ref*{tab:results-weibo16-deduplication}, we can see that if we train and validate the detectors on the deduplicated version of the dataset, the performances of the two detectors will increase on the original testing set (shown in bold in the table). Therefore, it verifies that training on the deduplicated datasets will enhance the generalization ability of the models to some extent. Moreover, if we fix the training and validation set deduplicated and just change the testing set from the original version to the deduplicated version, on \textbf{BiGRU} the macro F1 score and accuracy increase, while on \textbf{HSA-BLSTM} the metrics both decrease. We suppose the reasons are that on the original testing set, the detectors will predict the duplicated news pieces as highly similar results. So some clusters of duplicated pieces may be all predicted correctly, while others may be all predicted mistakenly, resulting in the unstable performance of the detectors. In a conclusion, deduplicating the dataset can help mitigate this issue.

\section*{Appendix B. The method to calculate the Dual Emotion Category}

It is mentioned in Section~\ref{sec:analysis} that we use the pretrained emotion classifiers to calculate the value of \textit{Dual Emotion Category}. 
The method to calculate the \textit{Dual Emotion Category} are as follows:

For publisher emotion, we feed the text of the news content into the emotion classifier and take the emotion with the maximum probability as the publisher emotion category. For social emotion, we feed the news comments once a time. After getting the output vector of each comment, each dimension of which represents the probability of the given comment having a certain kind of emotion, we average the probability vector of all the comments in each dimension. Finally, we take the emotion with the maximum probability as the social emotion category (i.e., soft voting). 

For example, assume that the the output of an emotion classifier is a probability vector on \textit{angry}, \textit{disgusting}, \textit{happy} and \textit{none} and the given news piece has two comments. The content probabilities are $[0.3, 0.1, 0, 0.6]$. So we can use the corresponding emotion of $0.6$, \textit{none}, as the publisher emotion category. The probability vector is $[0.8, 0.1, 0, 0.1]$ for the first comment, and $[0.6, 0.3, 0.1, 0]$ for the second comment. So we firstly average all the comment probability values and get $[0.7, 0.2, 0.05, 0.05]$. Then we use the corresponding emotion of $0.7$, \textit{angry}, as the news social emotion category. Thus, the categorical variable \textit{Dual Emotion Category} is none for publisher emotion and angry for social emotion.

\end{document}